\documentclass[10pt,twocolumn,letterpaper]{article}
\usepackage[pagenumbers]{cvpr}
\usepackage[T1]{fontenc}
\usepackage{graphicx}
\usepackage{booktabs}
\usepackage{array}
\usepackage{amsmath}
\usepackage{amssymb}
\usepackage{microtype}
\usepackage{multirow}
\usepackage{makecell}
\usepackage{xspace}
\usepackage{placeins}

\usepackage{pifont}
\newcommand{\tabyes}{\ding{51}}
\newcommand{\tabno}{\ding{55}}

\usepackage{float}
\newfloat{algorithm}{tbp}{loa}
\floatname{algorithm}{Algorithm}

\DeclareMathSizes{8.2}{8.2}{6}{5}
\DeclareMathSizes{7.8}{7.8}{6}{5}

\DeclareMathSizes{8.5}{8}{6}{5}

\usepackage{graphicx}

\definecolor{cvprblue}{rgb}{0.21,0.49,0.74}
\usepackage[pagebackref,breaklinks,colorlinks,allcolors=cvprblue]{hyperref}
\hypersetup{pdftitle={ReView: Online Zero-Shot 3D Instance Segmentation by Revisiting Past Views},pdfauthor={Yuanyang Zhang, Li Yao, Yijie Lin, Xinchao Wang}}

\title{ReView: Online Zero-Shot 3D Instance Segmentation\\by Revisiting Past Views}
\author{
Yuanyang Zhang\textsuperscript{1}\qquad
Li Yao\textsuperscript{1}\qquad
Yijie Lin\textsuperscript{2}\qquad
Xinchao Wang\textsuperscript{3}\\[4pt]
\textsuperscript{1}School of Computer Science and Engineering, Southeast University, Nanjing, China\\
\textsuperscript{2}The Hong Kong University of Science and Technology, Hong Kong, China\\
\textsuperscript{3}National University of Singapore, Singapore\\[2pt]
{\tt\small \{zhangyuanyang, yao.li\}@seu.edu.cn\quad linyijie.gm@gmail.com\quad xinchao@nus.edu.sg}
}

\begin{document}
\raggedbottom
\clubpenalty=10000
\widowpenalty=10000
\maketitle

\begin{abstract}
Online zero-shot 3D instance segmentation identifies objects during
scanning without training on 3D instance annotations. Existing methods
incrementally associate 2D masks to build 3D instances. However,
segmentation errors and missed instances can persist as observations
accumulate. Moreover, some methods assign identical confidence scores
to all instances, leaving reliable and unreliable instances
indistinguishable. We observe that masks from different past viewpoints
can reveal errors in accumulated instances, while their agreement with
an instance indicates its reliability. We therefore propose
\emph{ReView}, a training-free framework that revisits accumulated
instances at query time using only observations collected so far,
without changing the online association process. For each instance,
ReView finds the best-matching 2D mask in each valid past view. In
\emph{view-consensus refinement}, these masks provide votes on visible
points from views with sufficient agreement, removing wrongly included
points and adding missing ones before connectivity cleanup. Reusing
the same matches, ReView averages agreement across all valid views to
obtain an \emph{observation-consistency confidence} for ranking instances.
To recover missed objects, \emph{track-based recovery} adds candidates
from temporal mask tracks after overlap checks and duplicate suppression.
Experiments on ScanNet200 and SceneNN show that ReView achieves the
highest AP among the compared online and offline methods.
Code is available at \url{https://github.com/zhangyuanyang21/ReView-3DIS}.
\end{abstract}

\begin{figure}[t]
  \centering
  \includegraphics[width=\columnwidth]{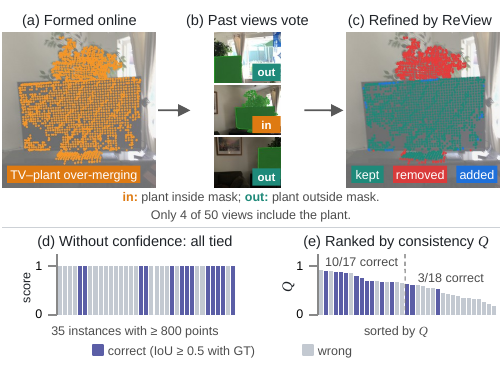}
  \caption{\textbf{ReView uses past views to refine and rank online instances.}
  (a)~The TV instance (orange) incorrectly includes points from the plant.
  (b)~Of the 50 voting views that see the plant, only 4 include it in their
  best-matching 2D masks.
  (c)~View-consensus refinement removes the wrongly included points (red)
  and adds missing points (blue).
  (d)~The 35 displayed instances, each with at least 800 points, initially
  have identical scores.
  (e)~Ranking by observation-consistency confidence $Q$ places 10 of the
  13 correct instances (indigo, IoU $\geq0.5$ with ground truth) among
  the top 17. Ground truth is used only to label the bars.}
  \label{fig:teaser}
\end{figure}

\section{Introduction}
3D instance segmentation identifies individual objects in a 3D scene and
the points belonging to each, supporting applications in robotics and
augmented reality (AR)~\cite{mccormac2018fusionpp,jatavallabhula2023conceptfusion}.
Supervised methods~\cite{jiang2020pointgroup,vu2022softgroup,schult2023mask3d}
perform well on predefined categories but require costly 3D instance
annotations and often struggle to generalize to unseen categories.
Zero-shot methods~\cite{lu2023ovir3d,yin2024sai3d,yan2024maskclustering}
instead use class-agnostic masks from 2D segmentation foundation
models~\cite{kirillov2023sam,qi2023cropformer}. They lift these masks into
3D and combine them across views to form instances without training on
3D instance annotations.

However, many zero-shot methods operate offline on complete
scans~\cite{zhao2025sam2object,nguyen2025any3dis,zhao2026mv3dis}.
Embodied applications instead need instance segmentation during an RGB-D
scan, when only part of the scene has been
observed~\cite{mccormac2018fusionpp,jatavallabhula2023conceptfusion}.
To meet this need, online zero-shot
methods~\cite{yang2023sam3d,tang2025onlineanyseg,xu2026stream3d} build
3D instances incrementally through an \emph{associate-and-accumulate}
process. They lift 2D masks from incoming frames or local windows into
3D and use spatial overlap or feature similarity to merge them with
existing instances or form new ones.

However, existing methods face two problems. (i)~Segmentation errors and
missed instances: A mask that crosses an object boundary can add points
from a neighbouring object (Fig.~\ref{fig:teaser}a). Conversely, if a 2D mask
misses part of an object, the resulting 3D instance may also miss that part.
Beyond these errors within an instance, an entire object may fail to form a
separate instance if its masks are discarded or merged into neighbouring
instances. These errors can persist as observations accumulate.
(ii)~Uninformative instance scores: Some methods assign the same
confidence score to all instances (Fig.~\ref{fig:teaser}d), so reliable
and unreliable instances are indistinguishable and average precision (AP)
evaluation cannot rank reliable predictions first.

We observe that past 2D masks can help identify parts that an instance
wrongly includes or misses (Fig.~\ref{fig:teaser}b,\,c). By query time, the
same object has often been observed from several viewpoints. The masks
from these views are generated by the 2D segmenter without relying on
previous 3D association decisions. Comparing them with the instance
therefore provides evidence for correcting its segmentation, while their
agreement offers a measure of its reliability. Motivated by this
observation, we propose \emph{ReView}, which revisits existing instances
using only observations available up to the query time, without changing
the online association process.

Specifically, ReView finds the best-matching 2D mask for each instance in
each valid past view and measures their agreement by IoU in 3D. In
\emph{view-consensus refinement}, views with sufficient agreement vote on
which points belong to the instance. These votes guide the removal of
wrongly included points and the addition of missing ones, followed by
connectivity cleanup. Reusing the same matches, the mean agreement across
all valid views provides an \emph{observation-consistency confidence} for
ranking instances (Fig.~\ref{fig:teaser}e). To recover objects missing from
the instance bank, \emph{track-based recovery} forms candidates from
temporal mask tracks and adds them after overlap checks against existing
instances and duplicate suppression among candidates. All steps are
training-free.

Experiments on ScanNet200 and SceneNN show that ReView achieves the highest
AP among the compared online and offline methods. With instances ranked by
observation-consistency confidence, ReView also outperforms the evaluated
supervised online method in full-input COCO-style AP$_{50}$~\cite{lin2014coco}
(C-AP$_{50}$). Our contributions are:
\begin{itemize}
  \item We propose a training-free framework for revisiting accumulated
  3D instances at query time. Using only observations available up to
  that time, it corrects segmentation errors and recovers missed objects
  without changing the online association process.
  \item We design view-consensus refinement and observation-consistency
  confidence based on the same multi-view mask matches. The matched
  2D masks provide point-wise votes to refine instance masks, while their
  agreement provides confidence scores for ranking instances by reliability.
  \item Experimental results show consistent AP gains across the evaluated
  scanning stages and improved C-AP$_{50}$ through confidence ranking on
  ScanNet200 and SceneNN. Refinement and connectivity cleanup also improve
  the full-input AP of three other online methods.
\end{itemize}

\section{Related Work}
\paragraph{Zero-shot 3D instance segmentation.}
Zero-shot 3D instance segmentation lifts class-agnostic masks from
2D models such as SAM~\cite{kirillov2023sam},
CropFormer~\cite{qi2023cropformer} and SAM~2~\cite{ravi2025sam2}
into 3D. Related approaches aggregate
superpoints~\cite{nguyen2024open3dis,yin2024sai3d},
propagate or track masks across
views~\cite{zhao2025sam2object,nguyen2025any3dis,kim2026cdis,zhou2025opentrack3d},
or refine 3D proposals~\cite{jung2025details,wang2026sgs3d}.
Among offline methods, MaskClustering~\cite{yan2024maskclustering}
groups 2D masks into 3D instances using view consensus, while
MV3DIS~\cite{zhao2026mv3dis} uses 3D-guided mask matching to refine
2D segmentation maps, then reassigns boundary superpoints using
affinities recomputed from these maps.
ReView instead revisits instances formed online at query time.
It uses matched masks from past views for point-wise voting and
reuses their agreement for confidence ranking, without requiring
superpoints or changing the online association process.

\paragraph{Online 3D instance segmentation.}
Online methods incrementally update 3D instances as new observations
arrive. Supervised approaches learn instance association and temporal
updates~\cite{xu2025embodiedsam,wang2025autoseg3d,wang2026samosaic3d},
while ESAM++~\cite{liu2026esampp} targets efficient deployment on
edge devices. Zero-shot methods avoid training on 3D instance
annotations. OnlineAnySeg~\cite{tang2025onlineanyseg} associates lifted
masks using spatial overlap, semantic and geometric cues.
Stream3D~\cite{xu2026stream3d} and Stream3Dv2~\cite{xu2026stream3dv2}
combine local multi-view segmentation with geometric refinement
and historical instance updates. Other methods address
open-vocabulary panoptic reconstruction~\cite{patel2025razer}
and monocular 3D segmentation~\cite{du2026moonseg3r}.
ReView adds a separate query-time stage that refines and ranks
accumulated instances without feeding changes back into online
association (Sec.~\ref{sec:plugin}).

\section{Method}
\label{sec:method}
\begin{figure*}[t]
  \centering
  \includegraphics[width=\textwidth]{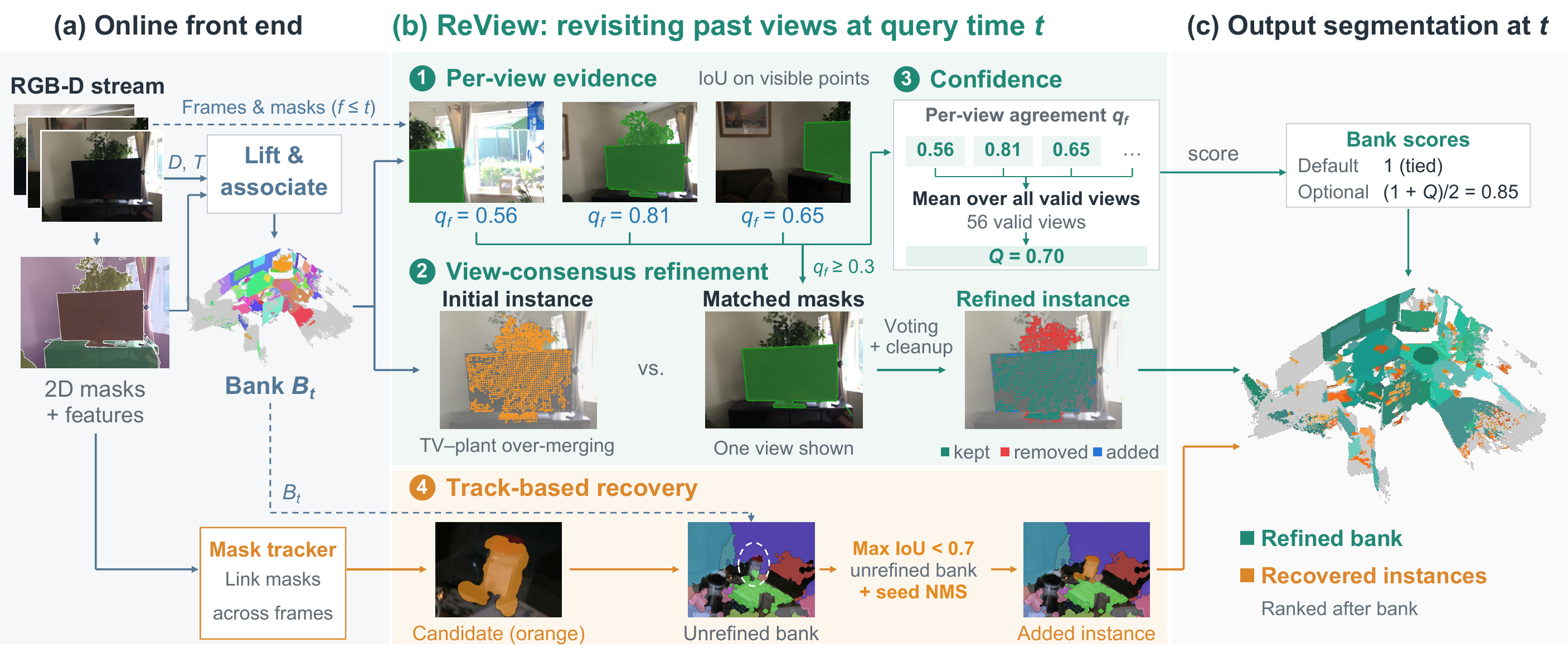}
  \caption{\textbf{Overview of ReView.}
  (a)~The online front end lifts and associates 2D masks to build
  the instance bank $\mathcal B_t$.
  (b)~At query time $t$, ReView uses only frames $f\leq t$ and their masks:
  (1)~best-matching masks in valid views provide IoU agreement $q_f(I)$
  over visible points;
  (2)~views with $q_f(I)\geq0.3$ vote point-wise to refine instances,
  followed by connectivity cleanup;
  (3)~mean agreement over all valid views gives $Q(I)$;
  (4)~online mask tracks yield recovery seeds, filtered by overlap
  checks (IoU $<0.7$ to every unrefined bank instance) and seed NMS.
  (c)~Refined bank instances score 1 by default or $(1+Q(I))/2$
  with optional confidence ranking. Recovered seeds $S_j$ score
  $0.499\,Q(S_j)$ and rank below all bank instances.}
  \label{fig:method}
\end{figure*}
\subsection{Problem Formulation}
\label{sec:formulation}
We consider online zero-shot 3D instance segmentation from posed RGB-D
frames $f=1,2,\dots$. A pretrained 2D segmenter
(CropFormer~\cite{qi2023cropformer}) produces non-overlapping,
class-agnostic masks for each frame. At query time $t$, only frames
$f\leq t$ and the scene $\mathcal P_t$ reconstructed from
them~\cite{niessner2013hashing} may be used.
The output is $\{(I_k,s_k)\}$, where $I_k\subseteq\mathcal P_t$
is the point set of instance $k$ and $s_k$ its confidence score.
The method requires no training on 3D instance annotations.

\subsection{Overview}
\label{sec:frontend}
The online front end lifts 2D masks into $\mathcal P_t$ and associates
them using spatial overlap, semantic features and geometric
descriptors~\cite{choy2019fcgf}, forming an instance bank $\mathcal B_t$
(Sec.~\ref{sec:protocol}).
As shown in Fig.~\ref{fig:method}, ReView revisits this bank at query
time $t$ using only archived frames $f\leq t$ and their 2D masks.
Per-view evidence matches each instance to masks in views where it
is visible (Sec.~\ref{sec:evidence}).
The matched masks provide point-wise votes for view-consensus
refinement, followed by connectivity cleanup (Sec.~\ref{sec:refine}).
Their agreement provides observation-consistency confidence
(Sec.~\ref{sec:confidence}).
Track-based recovery uses mask tracks maintained alongside the
online front end to generate additional candidates
(Sec.~\ref{sec:recovery}).
The refined bank and admitted candidates form the output
(Sec.~\ref{sec:output}).
Revisiting does not feed back into online association, allowing
refinement and confidence estimation to be applied to other
front ends (Sec.~\ref{sec:plugin}).

\subsection{Per-View Evidence}
\label{sec:evidence}
We compare each instance with past 2D masks over the points visible
in each view. For a frame $f\leq t$ with pose $T_f$ and depth map $D_f$,
a point of $\mathcal P_t$ is \emph{visible} if it projects inside the
image, has camera depth in $(0.1,5)$\,m, and differs from the depth
measurement at its projected pixel by at most 0.08\,m.
Let $V_f$ denote these visible points. Each 2D mask $\ell$ corresponds
to the lifted point set $M_{f,\ell}\subseteq V_f$, containing the
visible points whose projections fall inside the mask.

For an instance $I\in\mathcal B_t$, we use views with at least 50 visible
instance points,
$\mathcal G_I=\{f\leq t: |I\cap V_f|\geq 50\}$.
In each valid view, we select the mask with the highest IoU with the
visible part of $I$ and use this IoU as the agreement:
\begin{align}
 \ell^*_f(I)&=\arg\max\nolimits_\ell\ \mathrm{IoU}\big(I\cap V_f,\,M_{f,\ell}\big),\\
 q_f(I)&=\mathrm{IoU}\big(I\cap V_f,\,M_{f,\ell^*_f(I)}\big).
\end{align}
The masks are generated by the 2D segmenter without relying on the
front end's association decisions. The same matches and their
agreement $q_f(I)\in[0,1]$ support both refinement and confidence
estimation.

\subsection{View-Consensus Refinement}
\label{sec:refine}
We refine each instance by combining its matched masks from past
views (Fig.~\ref{fig:teaser}). For an instance $I$, only views with
sufficient agreement participate in voting:
$\mathcal F_I=\{f\in\mathcal G_I:q_f(I)\geq\tau_q\}$,
with $\tau_q=0.3$.
For each point $p\in\mathcal P_t$, we compute
\begin{equation}
 v_p=\sum_{f\in\mathcal F_I}\mathbf{1}[p\in V_f],\qquad
 u_p=\sum_{f\in\mathcal F_I}\mathbf{1}[p\in M_{f,\ell^*_f(I)}],
\end{equation}
where $\mathbf{1}[\cdot]$ is the indicator function, $v_p$ counts
the voting views that observe $p$, and $u_p$ counts those whose
matched masks contain it.
The refined instance is
\begin{equation}
 \hat I=\{p\in\mathcal P_t:\ v_p\geq n,\ u_p\geq\rho\,v_p\}
 \ \cup\ \{p\in I:\ v_p<n\},
 \label{eq:refine}
\end{equation}
with $n=2$ and $\rho=\tfrac12$.
A point observed by at least $n$ voting views is included if at
least half of those views' matched masks contain it; otherwise,
it is excluded. Points with fewer observations retain their
original membership. This rule can both remove wrongly included
points and add missing ones.

\paragraph{Connectivity cleanup.}
Matched masks may include neighbouring surfaces across several
views, leaving disconnected fragments after voting.
We build a $\delta$-neighbourhood graph on $\hat I$
($\delta=5$\,cm) and retain its largest connected component only
if it contains at least half of the points.
Otherwise, $\hat I$ remains unchanged.

\subsection{Observation-Consistency Confidence}
\label{sec:confidence}
To distinguish otherwise tied instances, we assess how well each
instance agrees with its past 2D observations.
For a bank instance $I$ before refinement, we define its confidence
as the mean agreement over all valid past views:
\begin{equation}
 Q(I)=\frac{1}{|\mathcal G_I|}
 \sum_{f\in\mathcal G_I}q_f(I).
\end{equation}
We set $Q(I)=0$ when $\mathcal G_I=\emptyset$.

The mean includes valid views with $q_f(I)<\tau_q$, retaining
evidence of disagreement that is excluded from refinement voting.
The same matches thus support both point-wise refinement and
instance-level scoring, without training a scoring head or
performing additional mask matching for bank instances.

\subsection{Track-Based Recovery}
\label{sec:recovery}
\paragraph{Mask tracking.}
To recover objects missing from the bank, a causal tracker links
lifted masks across frames alongside the online front end.
It uses mutual best matching based on voxel overlap and CLIP
similarity, subject to the matching criteria in the appendix.
Unmatched masks start new tracks.
At query time $t$, for each track $j$ with at least three masks,
we map the union of its mask voxels to $\mathcal P_t$.
We take the largest connected component (10\,cm neighbourhood)
as a seed $S_j$ if this component contains at least 50 points.
Each seed's confidence $Q(S_j)$ is the mean agreement over its own
valid views (Sec.~\ref{sec:confidence}).

\paragraph{Seed admission.}
To limit overlap with existing instances, a seed must satisfy
\begin{equation}
 \max\nolimits_{I\in\mathcal B_t}\mathrm{IoU}(S_j,I)<\tau_r,
\end{equation}
where $\tau_r=0.7$ and $\mathcal B_t$ is the bank before refinement.
We then apply greedy non-maximum suppression among the remaining
seeds at the same IoU threshold, using a priority score based on
track observation count and seed size (see the appendix).

\subsection{Output}
\label{sec:output}
At query time $t$, ReView outputs the refined and cleaned bank
instances together with the admitted recovery seeds.
By default, bank instances score 1 and seeds $S_j$ score
$0.499\,Q(S_j)\in[0,0.499]$.
Keeping the same masks and seed scores, \emph{ReView + conf.}
assigns each refined bank instance $\hat I$ the score
$(1+Q(I))/2\in[0.5,1]$, using $Q(I)$ computed before refinement.
In both settings, seeds rank below all bank instances and are
ordered by $Q$.

\section{Experiments}
\subsection{Setup}
\label{sec:protocol}
\paragraph{Datasets.}
We evaluate on ScanNet200~\cite{dai2017scannet,rozenberszki2022scannet200}
and SceneNN~\cite{hua2016scenenn}, two indoor RGB-D benchmarks with
instance annotations. ScanNet200 covers 200 semantic categories.
Following prior work~\cite{tang2025onlineanyseg,xu2025embodiedsam},
we use all 312 validation sequences.
For SceneNN, we use the same 12 scenes as
OnlineAnySeg~\cite{tang2025onlineanyseg}, with scene IDs listed in
the appendix. To assess performance during scanning, we evaluate
after processing 25\%, 50\%, 75\% and 100\% of the available input sequence.

\paragraph{Baselines.}
We compare with offline and online methods.
OVIR-3D~\cite{lu2023ovir3d} and
MaskClustering~\cite{yan2024maskclustering} perform offline zero-shot
segmentation using complete scans.
Among online methods, EmbodiedSAM~\cite{xu2025embodiedsam} learns
mask association from ScanNet200 annotations.
The other three methods are zero-shot:
SAM3D~\cite{yang2023sam3d} merges masks frame by frame;
OnlineAnySeg~\cite{tang2025onlineanyseg} combines spatial overlap
with semantic and geometric cues;
and Stream3D~\cite{xu2026stream3d} groups masks within local windows
and associates them with historical instances.
We rerun the online baselines using released implementations on
the same input sequences and poses under a common evaluation protocol.
SAM3D and EmbodiedSAM use SAM masks, while OnlineAnySeg and
Stream3D use CropFormer masks.
Frame-sampling settings are detailed in the appendix.
Since the released Stream3D implementation loads the complete
scene mesh, we report its results only at full input.

\paragraph{Implementation details.}
In the main experiments, we use the released OnlineAnySeg
front end~\cite{tang2025onlineanyseg} with
CropFormer masks~\cite{qi2023cropformer} and
CLIP ViT-H/14 features~\cite{radford2021clip}.
We process every 10th frame on ScanNet200 and every 20th frame
on SceneNN.
The front end and tracker run on one NVIDIA RTX 3090 GPU,
while query-time revisiting runs on one CPU core (AMD EPYC 7413).
On 20 ScanNet200 scans, online tracking and query-time revisiting
together add an amortized 21\,ms per processed frame (1.2\% of the
front-end runtime), with one query per scan.
A query takes 1.4\,s (median), excluding time spent loading
archived frames.
Detailed timings are provided in the appendix.

\begin{table*}[!t]
  \caption{\textbf{Full-input results.}
  Evaluation uses OnlineAnySeg's released script
  (Sec.~\ref{sec:protocol}). Online methods are our reruns on
  the same input sequences; ReView + conf.\ ranks the same masks by confidence
  (Sec.~\ref{sec:published}). $^{\ddagger}$: offline, reported
  in~\cite{tang2025onlineanyseg} with the same script. $^{\dagger}$: SAM3D without its optional over-segmentation ensemble. $^{\S}$: released code with our masks; uses the complete scene
  mesh and is included only as a full-input reference.
  \textbf{Bold}/\underline{underline}: best/second best per column.}
  \label{tab:main}
  \centering
  \small
  \setlength{\tabcolsep}{6pt}
  \renewcommand{\arraystretch}{1.08}
  \begin{tabular}{l|cc|ccc|ccc}
\specialrule{1pt}{0pt}{0pt}
& & & \multicolumn{3}{c|}{\textbf{ScanNet200}} & \multicolumn{3}{c}{\textbf{SceneNN}}\\
\textbf{Method} & \textbf{Online} & \textbf{Zero-shot} & AP & AP$_{50}$ & AP$_{25}$ & AP & AP$_{50}$ & AP$_{25}$\\
\specialrule{1pt}{0pt}{0pt}
OVIR-3D~\cite{lu2023ovir3d}$^{\ddagger}$ & \tabno & \tabyes & 14.4 & 27.5 & 38.8 & 12.3 & 24.4 & 34.6 \\
MaskClustering~\cite{yan2024maskclustering}$^{\ddagger}$ & \tabno & \tabyes & \underline{19.7} & 36.4 & 51.4 & 16.3 & 31.7 & 46.2 \\
\hline
EmbodiedSAM~\cite{xu2025embodiedsam} & \tabyes & \tabno & 14.80 & 36.57 & \textbf{64.43} & 11.18 & 30.99 & 61.48 \\
SAM3D~\cite{yang2023sam3d}$^{\dagger}$ & \tabyes & \tabyes & 9.63 & 23.75 & 49.85 & 10.05 & 23.26 & 47.95 \\
Stream3D~\cite{xu2026stream3d}$^{\S}$ & \tabyes & \tabyes & 19.18 & 34.31 & 53.44 & 16.27 & 30.43 & 47.25 \\
OnlineAnySeg~\cite{tang2025onlineanyseg} & \tabyes & \tabyes & 18.96 & 36.77 & 55.63 & 18.12 & 36.17 & 59.86 \\
\hline
\textbf{ReView} (ours) & \tabyes & \tabyes & \textbf{21.38} & \textbf{41.05} & 62.51 & \textbf{20.75} & \underline{39.33} & \textbf{62.43} \\
\textbf{ReView} + conf. (ours) & \tabyes & \tabyes & 19.34 & \underline{40.31} & \underline{63.12} & \underline{18.97} & \textbf{39.82} & \underline{62.06} \\
\specialrule{1pt}{0pt}{0pt}
\end{tabular}

\end{table*}

\begin{table*}[!t]
  \caption{\textbf{Segmentation results during scanning.}
  Percentages indicate the fraction of the available input sequence
  processed.
  Each stage uses only observations available so far and is evaluated
  against the complete scene ground truth.
  \textbf{Bold}/\underline{underline}: best/second best per column
  within each dataset.}
  \label{tab:prefix}
  \centering
  \footnotesize
  \setlength{\tabcolsep}{3.0pt}
  \renewcommand{\arraystretch}{1.08}
  \begin{tabular}{ll|ccc|ccc|ccc|ccc}
\specialrule{1pt}{0pt}{0pt}
& & \multicolumn{3}{c|}{\textbf{25\%}} & \multicolumn{3}{c|}{\textbf{50\%}} & \multicolumn{3}{c|}{\textbf{75\%}} & \multicolumn{3}{c}{\textbf{100\%}}\\
\textbf{Data} & \textbf{Method} & AP & AP$_{50}$ & C-AP$_{50}$ & AP & AP$_{50}$ & C-AP$_{50}$ & AP & AP$_{50}$ & C-AP$_{50}$ & AP & AP$_{50}$ & C-AP$_{50}$\\
\specialrule{1pt}{0pt}{0pt}
\multirow{5}{*}{ScanNet200} & EmbodiedSAM & \underline{6.74} & \textbf{17.25} & \textbf{17.73} & 10.93 & \textbf{27.39} & \textbf{27.65} & 13.55 & 33.81 & \underline{34.16} & 14.80 & 36.57 & \underline{36.87} \\
 & SAM3D & 4.33 & 11.23 & 4.24 & 6.92 & 17.56 & 6.70 & 8.83 & 22.12 & 8.63 & 9.63 & 23.75 & 9.36 \\
 & OnlineAnySeg & 6.70 & 14.16 & 8.64 & \underline{11.98} & 24.65 & 15.59 & \underline{16.26} & 32.27 & 21.22 & 18.96 & 36.77 & 24.50 \\
 & \textbf{ReView} (ours) & \textbf{7.63} & \underline{16.24} & 10.82 & \textbf{13.40} & \underline{27.32} & 18.42 & \textbf{18.18} & \textbf{35.97} & 24.98 & \textbf{21.38} & \textbf{41.05} & 29.08 \\
 & ReView + conf. (ours) & 5.12 & 13.05 & \underline{13.85} & 10.30 & 24.28 & \underline{24.60} & 15.49 & \underline{34.02} & \textbf{34.40} & \underline{19.34} & \underline{40.31} & \textbf{40.64} \\
\hline
\multirow{5}{*}{SceneNN} & EmbodiedSAM & 2.82 & 8.50 & \textbf{9.15} & 4.95 & 15.18 & \underline{15.90} & 9.26 & 26.95 & \underline{27.29} & 11.18 & 30.99 & \underline{31.36} \\
 & SAM3D & 3.06 & 7.51 & 2.10 & 5.41 & 12.31 & 3.57 & 8.34 & 20.13 & 6.65 & 10.05 & 23.26 & 7.97 \\
 & OnlineAnySeg & \underline{4.58} & \underline{9.29} & 4.58 & \underline{9.23} & \textbf{20.43} & 11.35 & \underline{13.60} & \underline{29.34} & 16.87 & 18.12 & 36.17 & 21.52 \\
 & \textbf{ReView} (ours) & \textbf{5.20} & \textbf{9.78} & 5.07 & \textbf{10.20} & \underline{20.08} & 11.19 & \textbf{15.61} & \textbf{31.96} & 19.29 & \textbf{20.75} & \underline{39.33} & 24.56 \\
 & ReView + conf. (ours) & 3.33 & 6.99 & \underline{7.83} & 6.98 & 15.74 & \textbf{16.79} & 11.93 & 28.39 & \textbf{29.48} & \underline{18.97} & \textbf{39.82} & \textbf{40.59} \\
\specialrule{1pt}{0pt}{0pt}
\end{tabular}

\end{table*}

\paragraph{Metrics.}
We evaluate class-agnostic 3D instance segmentation using the
evaluation script released by OnlineAnySeg~\cite{tang2025onlineanyseg}.
We report AP$_{25}$ and AP$_{50}$ at IoU thresholds of 25\% and
50\%, respectively, and AP averaged over thresholds from 50\%
to 90\% in steps of 5\%.
Predicted masks are mapped to the ground-truth point cloud using
nearest-neighbour correspondences from GT points to reconstruction
points covered by predictions.
AP is computed by pooling predictions across scenes.
Table values are percentages, with percent signs omitted.
Detailed mapping rules and results under fixed correspondence are
provided in the appendix.

\begin{figure}[t]
  \centering
  \includegraphics[width=\columnwidth]{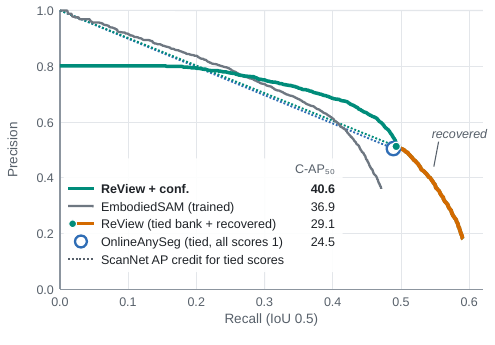}
  \caption{\textbf{Pooled precision--recall at IoU 0.5 on
  ScanNet200 (full input).}
  The hollow and filled circles mark OnlineAnySeg's tied predictions
  and ReView's tied bank, respectively.
  Dotted lines connect these points to $(0,1)$ for ScanNet AP
  integration.
  The teal curve ranks bank instances by confidence;
  the orange recovery tail is shared by both ReView variants.
  Curves show interpolated precision, and legend values are
  C-AP$_{50}$.}
  \label{fig:pr}
\end{figure}

\paragraph{AP for tied predictions.}
\label{sec:analysis}
At a fixed IoU threshold, predictions with identical confidence
scores produce a single precision--recall point $(R,P)$, where
$R$ and $P$ denote recall and precision.
The ScanNet evaluator connects this point to $(0,1)$ and applies
trapezoidal integration, giving $R(1+P)/2$.
Confidence ranking introduces intermediate points and can lower
this area even when the masks and final $(R,P)$ are unchanged
(Fig.~\ref{fig:pr}).
To complement this metric, we report COCO-style
AP$_{50}$~\cite{lin2014coco} (C-AP$_{50}$), which averages
interpolated precision at 101 equally spaced recall levels
from 0 to 1.
We retain the same instance matches and group tied scores together,
changing only the integration rule.
Derivations are provided in the appendix.

\begin{figure}[t]
  \centering
  \includegraphics[width=\columnwidth]{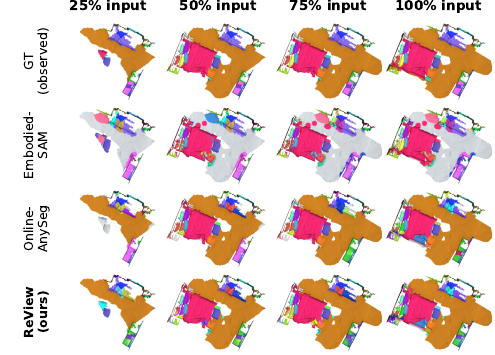}
  \caption{\textbf{Segmentation during scanning} on a ScanNet200
  scene at 25--100\% input (top view).
  Predictions are displayed on the observed portion of the GT mesh,
  with matched instances sharing GT colours and unlabelled
  surfaces shown in grey.
  The GT row is restricted to the shared front end's reconstruction
  for visualization only.}
  \label{fig:online}
\end{figure}

\begin{figure*}[t]
  \centering
  \includegraphics[width=\textwidth]{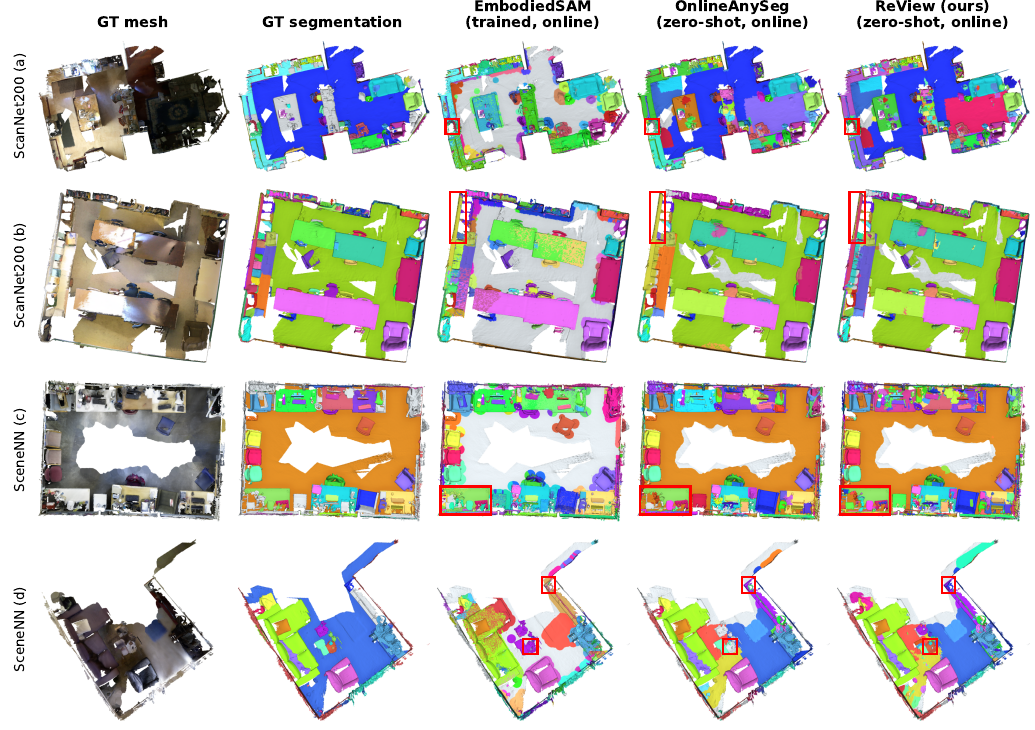}
  \caption{\textbf{Full-input segmentation comparison} on ScanNet200
  and SceneNN scenes (top view, ceiling removed).
  Predictions are mapped to the GT mesh.
  Matched predictions share GT colours (Hungarian matching,
  IoU $\geq0.25$); grey denotes unlabelled surface.
  Red boxes highlight objects that ReView segments more accurately
  than either baseline.}
  \label{fig:scenes}
\end{figure*}

\subsection{Quantitative Results}
\label{sec:published}
\paragraph{Full-input results.}
Table~\ref{tab:main} compares full-input segmentation results
on ScanNet200 and SceneNN.
ReView achieves the highest AP among the compared methods
on both datasets.
Compared with the offline zero-shot method MaskClustering,
it improves AP by 1.7 and 4.5 points, respectively.
Among online zero-shot methods, ReView outperforms SAM3D
and Stream3D.
Compared with OnlineAnySeg, it improves AP by 2.4 and 2.6 points
and AP$_{50}$ by 4.3 and 3.2 points, respectively.
ReView also achieves higher AP than the supervised baseline
EmbodiedSAM on both datasets, without training on 3D instance
annotations.
These gains show that revisiting accumulated instances using
past observations improves segmentation without changing
the online association process.

\paragraph{Intermediate-input results.}
To evaluate segmentation during scanning, we compare methods
after processing 25\%, 50\%, 75\% and 100\% of the available
input sequence, using only observations available at each stage
(Table~\ref{tab:prefix}).
ReView achieves the highest AP at all four stages on both datasets.
Compared with OnlineAnySeg, its AP gains increase from 0.9 to
2.4 points on ScanNet200 and from 0.6 to 2.6 points on SceneNN
as the input grows from 25\% to 100\%.
These results show that ReView improves segmentation during
scanning, with larger gains at later evaluated stages.

\paragraph{Confidence ranking.}
To assess the ranking effect of observation-consistency confidence,
we compare ReView with ReView + conf.\ (Fig.~\ref{fig:pr} and
Table~\ref{tab:prefix}). Only the scores of bank instances change;
all instance masks and the scores of recovered instances remain fixed.
Figure~\ref{fig:pr} shows the full-input precision--recall results
on ScanNet200. Confidence ranking turns the tied bank's single
operating point into a curve, with higher precision than EmbodiedSAM
at medium recall.
With masks unchanged, full-input C-AP$_{50}$ increases from 29.1
to 40.6 on ScanNet200 and from 24.6 to 40.6 on SceneNN,
showing that observation consistency provides a more effective
instance ranking.
Table~\ref{tab:prefix} further shows that ranking improves
C-AP$_{50}$ at all four evaluated stages, achieving the highest
values among the listed methods at 75\% and 100\% input on both
datasets.
The decrease in ScanNet AP after ranking is consistent with
the integration effect for tied predictions discussed in
Sec.~\ref{sec:analysis}.

\subsection{Qualitative Results}
\label{sec:qualitative}
\paragraph{During scanning.}
Figure~\ref{fig:online} compares ReView, EmbodiedSAM and OnlineAnySeg
on the same ScanNet200 scene at four input stages.
At 25\% input, ReView assigns instance labels to observed object
regions that OnlineAnySeg leaves unlabelled. This coverage advantage
remains visible as more of the scene is observed.
Compared with EmbodiedSAM, ReView also shows more coherent segmentation
of the bed and nearby regions in the intermediate outputs.
These experimental results show that revisiting improves segmentation
before the scan is complete.

\paragraph{Full-scene comparison.}
Figure~\ref{fig:scenes} compares full-input segmentation results
of ReView, EmbodiedSAM and OnlineAnySeg on ScanNet200 and
SceneNN scenes.
Compared with OnlineAnySeg, ReView more completely segments the
boxed objects that are missed or incorrectly merged with neighbouring
regions. For three boxed objects in panels (a) and (d), OnlineAnySeg's
IoU with GT is at most 0.09, whereas ReView achieves 0.53--1.00.
Compared with EmbodiedSAM, ReView separates the boxed objects more
clearly from their surroundings, particularly in panels (c) and (d).
These experimental results show that ReView recovers missed object
regions and improves local segmentation at full input.

\subsection{Ablation Study}
\label{sec:ablation}
Table~\ref{tab:ablation} evaluates component removal, recovery candidate
ranking and recovery admission configurations, using ReView in
Table~\ref{tab:main} as the full method.
All variants keep bank instance scores at 1.
Confidence ranking of bank instances is evaluated separately in
Sec.~\ref{sec:published}.

\begin{table}[!h]
  \caption{\textbf{Ablations and recovery controls at full input.}
  \emph{w/o} denotes removal of the specified component;
  \emph{w/o $Q$-based seed ranking} ranks recovery seeds by native
  track scores instead of $Q$. Bold: best per column.}
  \label{tab:ablation}
  \centering
  \small
  \setlength{\tabcolsep}{4pt}
  \renewcommand{\arraystretch}{1.08}
  \begin{tabular}{l|cc|cc}
\specialrule{1pt}{0pt}{0pt}
\multirow{2}{*}{\textbf{Variant}} & \multicolumn{2}{c|}{\textbf{ScanNet200}} & \multicolumn{2}{c}{\textbf{SceneNN}}\\
& AP & AP$_{50}$ & AP & AP$_{50}$\\
\specialrule{1pt}{0pt}{0pt}
\emph{w/o} view-consensus refinement & 20.00 & 40.27 & 19.80 & 39.83 \\
\emph{w/o} connectivity cleanup & 19.76 & 39.80 & 19.61 & 37.74 \\
\emph{w/o} refinement \& cleanup & 20.35 & 40.71 & 19.76 & \textbf{40.65} \\
\emph{w/o} track recovery & 20.36 & 38.06 & 19.65 & 36.68 \\
\emph{w/o} $Q$-based seed ranking & 21.18 & 40.59 & 20.66 & 39.18 \\
containment-rule recovery & \textbf{21.39} & 40.17 & \textbf{20.89} & 38.19 \\
\hline
\textbf{Full method} & 21.38 & \textbf{41.05} & 20.75 & 39.33 \\
\specialrule{1pt}{0pt}{0pt}
\end{tabular}

\end{table}

\paragraph{Refinement and cleanup.}
View-consensus refinement re-estimates instance point sets from past-view
votes, while connectivity cleanup removes disconnected fragments.
Removing either component lowers AP by 1.4--1.6 points on ScanNet200
and about 1 point on SceneNN.
Using both yields higher AP than retaining either alone, supporting
their complementary roles.
On SceneNN, the full method has higher AP but lower AP$_{50}$ than the
variant without both components, indicating greater gains at stricter
IoU thresholds.

\paragraph{Track-based recovery.}
Removing recovery causes the largest AP$_{50}$ drop, by 3.0 points on
ScanNet200 and 2.7 on SceneNN.
Figure~\ref{fig:pr} further shows that adding recovery candidates raises
recall at IoU 0.5 from 0.49 to 0.59 on ScanNet200, supporting their role
in recovering missed instances.
With the candidate set unchanged, ranking by observation consistency
instead of native track scores further improves AP by 0.2 and 0.1 points
on the two datasets, respectively.

\paragraph{Recovery configuration.}
We also compare a configuration combining containment-based admission
with native track scores. A candidate is admitted only if its overlap
with every bank instance covers less than half of the candidate's points.
The default combination of IoU-based admission and observation-consistency
ranking achieves similar AP, but higher AP$_{50}$ by 0.9 points on
ScanNet200 and 1.1 on SceneNN.
This comparison reflects the joint effect of admission and ranking.

\begin{figure}[t]
  \centering
  \includegraphics[width=\columnwidth]{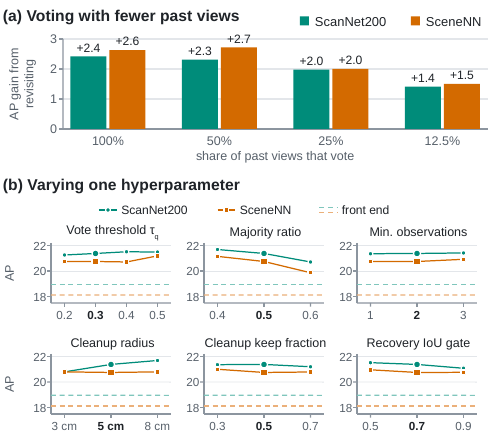}
  \caption{\textbf{Sensitivity analysis at full input.}
  (a)~AP gains over the front end when varying the fraction of past
  views used for voting and confidence, with front-end inputs unchanged.
  (b)~AP when varying one hyperparameter while keeping the others
  at their defaults.
  Bold tick labels and larger markers indicate default settings;
  dashed lines show front-end AP.}
  \label{fig:sensitivity}
\end{figure}

\subsection{Hyperparameter Sensitivity Analysis}
\label{sec:sensitivity}
At full input, we vary the share of past views used for voting and
confidence, keeping front-end inputs unchanged (Fig.~\ref{fig:sensitivity}a).
With 50\% of past views, AP differs from the defaults by about 0.1 on
both datasets. At 12.5\% (every eighth processed frame), gains over the
front end remain 1.4 and 1.5 AP on ScanNet200 and SceneNN, respectively,
retaining over half the default gains. ReView thus retains segmentation
gains with fewer revisited views.

We then vary one hyperparameter at a time, keeping the others at their
defaults (Fig.~\ref{fig:sensitivity}b). All tested settings outperform the
front end. The vote threshold $\tau_q$ selects views by mask agreement,
whereas the majority ratio $\rho$ sets the fraction of supporting votes
needed to include a point. Varying $\tau_q$, minimum observations,
cleanup keep fraction, or recovery IoU gate changes AP by less than 0.5
from the defaults on both datasets. Raising $\rho$ to 0.6 lowers AP by
0.7--0.9, so stricter voting does not necessarily improve segmentation.
Reducing the cleanup radius to 3\,cm lowers AP on ScanNet200 but leaves
SceneNN nearly unchanged. Full results and parameter selection details
are in the appendix.

\begin{table}[t]
  \caption{\textbf{ReView components applied to other methods}
  (ScanNet200, full input). Refinement uses the same CropFormer masks
  and hyperparameters as ReView. Original instance scores are retained
  unless conf.\ (confidence ranking) is added.
  $^{\dagger}$: SAM3D without its optional over-segmentation ensemble; $^{\S}$: uses the complete scene mesh.
  Bold: best within each method.}
  \label{tab:plugin}
  \centering
  \footnotesize
  \setlength{\tabcolsep}{3.5pt}
  \begin{tabular}{l|ccc|c}
\specialrule{1pt}{0pt}{0pt}
\textbf{Method} & AP & AP$_{50}$ & AP$_{25}$ & C-AP$_{50}$\\
\specialrule{1pt}{0pt}{0pt}
SAM3D$^{\dagger}$ & 9.63 & 23.75 & 49.85 & 9.36 \\
\quad + refinement & 14.93 & 30.14 & 51.97 & 13.53 \\
\quad + refinement + cleanup & \textbf{16.81} & 31.60 & 52.28 & 14.85 \\
\quad + refinement + cleanup + conf. & 15.24 & \textbf{32.52} & \textbf{56.61} & \textbf{33.11} \\
\hline
Stream3D$^{\S}$ & 19.18 & 34.31 & 53.44 & 15.51 \\
\quad + refinement & 17.60 & 34.56 & 53.75 & 15.92 \\
\quad + refinement + cleanup & \textbf{20.96} & \textbf{36.92} & 54.33 & 17.53 \\
\quad + refinement + cleanup + conf. & 18.88 & 36.59 & \textbf{55.62} & \textbf{36.94} \\
\hline
EmbodiedSAM & 14.80 & 36.57 & 64.43 & 36.87 \\
\quad + refinement & 8.40 & 28.24 & 63.01 & 28.45 \\
\quad + refinement + cleanup & \textbf{16.67} & \textbf{40.00} & \textbf{64.54} & \textbf{40.14} \\
\specialrule{1pt}{0pt}{0pt}
\end{tabular}

\end{table}

\subsection{Generalization to Other Base Methods}
\label{sec:plugin}
To test whether ReView's refinement and connectivity cleanup generalize
across online front ends, we apply them to SAM3D, Stream3D and
EmbodiedSAM on ScanNet200 (Table~\ref{tab:plugin}).
At full input, AP increases by 7.2, 1.8 and 1.9 points, respectively.
Refinement alone lowers AP for Stream3D and EmbodiedSAM, while adding
cleanup brings both above their baselines.
Refinement and cleanup improve SAM3D's AP at all four stages.
For EmbodiedSAM, they improve AP at 75--100\% input, leave it nearly
unchanged at 50\%, and lower it at 25\% (see the appendix).
These results show that refinement and connectivity cleanup together
improve full-input segmentation across different base methods.

Keeping the refined and cleaned instance masks fixed,
observation-consistency confidence ranking further raises
C-AP$_{50}$ from 14.9 to 33.1 for SAM3D and from 17.5 to 36.9
for Stream3D.
These gains show that observation consistency provides a more effective
ranking of the same instance predictions.

\section{Conclusion}
We presented ReView, a training-free method that improves online zero-shot
3D instance segmentation by revisiting past views. At query time, ReView
matches existing instances to past 2D masks and uses multi-view voting to
remove wrongly included points and add missing ones. The agreement of
these same matches provides confidence for ranking predictions by
reliability. Temporal mask tracks further recover objects that have not
formed separate instances. The entire process uses only observations
available up to the query time and leaves the online association process
unchanged. On ScanNet200 and SceneNN, ReView achieves the highest AP among
the compared methods. Its refinement and connectivity cleanup also improve
the full-input AP of three other online methods.

\paragraph{Limitations.}
ReView refines instances through consensus among the underlying
2D masks. When these masks share similar errors across views,
the erroneous regions may receive consistent support and remain
after refinement. Future work will explore how to reduce the
influence of such correlated mask errors.

{
\small
\setlength{\bibsep}{1pt}
\bibliographystyle{ieeenat_fullname}
\bibliography{references}
}

\clearpage
\appendix
\renewcommand{\thesection}{\Alph{section}}
\setcounter{table}{0}\renewcommand{\thetable}{S\arabic{table}}
\setcounter{figure}{0}\renewcommand{\thefigure}{S\arabic{figure}}
\setcounter{equation}{0}\renewcommand{\theequation}{S\arabic{equation}}
\makeatletter\renewcommand{\thealgorithm}{S\arabic{algorithm}}\makeatother
\twocolumn[{\centering\Large\textbf{\thetitle}\\
  \vspace{0.5em}Appendix\\
  \vspace{1.0em}}]

\noindent This appendix gives the remaining method details
(Sec.~\ref{sec:supp-method}), the evaluation protocol
(Sec.~\ref{sec:supp-evaluation}), the baseline configurations
(Sec.~\ref{sec:supp-baselines}), additional results
(Sec.~\ref{sec:supp-results}) and qualitative examples
(Sec.~\ref{sec:supp-qual}).

\section{Method Details}
\label{sec:supp-method}
\paragraph{Pseudocode.} Algorithm~\ref{alg:review} summarizes query-time
refinement, confidence scoring and recovery.

\begin{algorithm}[H]
\caption{Revisiting stage of ReView at query time $t$ (main text, Sec.~3). Hyperparameters: $\tau_q=0.3$, $n=2$, $\rho=\tfrac12$, cleanup radius 0.05\,m, $\tau_r=0.7$. For a bank instance or seed $X$, $\mathcal G_X$ contains archived views with at least 50 visible points of $X$.}
\label{alg:review}
\footnotesize
\begin{tabular}{@{}r@{\hspace{5pt}}p{0.89\columnwidth}@{}}
\toprule
\multicolumn{2}{@{}p{0.97\columnwidth}@{}}{\textbf{Input:} bank $\mathcal B_t$ on $\mathcal P_t$; posed RGB-D frames $f\leq t$ with 2D masks; track seeds $\{S_j\}$.}\\
\midrule
1: & $v_{I,p}\gets0$, $u_{I,p}\gets0$ for each $I\in\mathcal B_t$, $p\in\mathcal P_t$\\
2: & \textbf{for} each archived frame $f\leq t$ \textbf{do}\\
3: & \quad $V_f\gets$ visible points; lift each 2D mask to $M_{f,\ell}\subseteq V_f$\\
4: & \quad \textbf{for} each bank instance or seed $X$ with $f\in\mathcal G_X$ \textbf{do}\\
5: & \qquad match $X\cap V_f$ to $M_{f,\ell}$; record the highest IoU as $q_f(X)$ (0 if no mask exists)\\
6: & \qquad \textbf{if} $X$ is a bank instance $I$ and $q_f(I)\geq\tau_q$:\\
7: & \qquad\quad increment $v_{I,p}$ on $V_f$ and $u_{I,p}$ on the matched mask\\
8: & \textbf{for} each $I\in\mathcal B_t$ \textbf{do}\\
9: & \quad $\hat I\gets\{p\in\mathcal P_t:v_{I,p}\geq n,\,u_{I,p}\geq\rho v_{I,p}\}$\newline
\hspace*{1em}$\phantom{\hat I\gets{}}\cup\{p\in I:v_{I,p}<n\}$\\
10: & \quad keep the largest connected component of $\hat I$ if it holds $\geq$ half of $\hat I$\\
11: & for each bank instance or seed $X$, average $q_f(X)$ over $\mathcal G_X$ to obtain $Q(X)$; use 0 if $\mathcal G_X=\emptyset$\\
12: & admit seeds with IoU $<\tau_r$ to every unrefined bank instance; apply seed NMS in track-priority order\\
13: & score each $\hat I$ by 1 (default) or $(1+Q(I))/2$ (optional); score each admitted $S_j$ by $0.499\,Q(S_j)$\\
14: & \textbf{return} the scored bank instances and admitted seeds\\
\bottomrule
\end{tabular}
\end{algorithm}

\paragraph{Temporal tracker for recovery.} The tracker links masks causally
using the front end's masks and CLIP features, without changing the bank.
Each lifted mask is represented by its voxel set $V$ and unit-normalized
CLIP feature $\mathbf z$. Track $j$ stores the voxel sets $\mathcal H_j$
of its most recent four masks (or fewer for a new track) and a
unit-normalized feature prototype $\bar{\mathbf z}_j$. For a mask--track
pair, we define
\begin{align}
 O(V,j)&=\max_{H\in\mathcal H_j}\frac{|V\cap H|}{\min(|V|,|H|)},\\
 J(V,j)&=\max_{H\in\mathcal H_j}\frac{|V\cap H|}{|V\cup H|},\\
 A(V,j)&=\langle\mathbf z,\bar{\mathbf z}_j\rangle,\\
 w(V,j)&=0.7\,O(V,j)+0.3\,A(V,j).
\end{align}
Here, $O$ and $J$ measure spatial overlap, $A$ measures feature similarity,
and $w$ ranks candidate matches.

Only tracks observed within the preceding 30 frame indices are considered.
A candidate pair must satisfy $O\geq0.12$, $J\geq0.02$, $A\geq0.60$ and
$w\geq0.30$. Association requires mutual best matches under $w$, with the
best score exceeding the second-best by at least 0.04 in both directions.
An absent second-best candidate has score 0. When $O\geq0.60$, only the
margin requirement is waived.

A successful match updates the mask history, increments the observation
count $h_j$, and updates the feature prototype:
\[
 \bar{\mathbf z}_j\leftarrow
 \operatorname{normalize}(0.8\bar{\mathbf z}_j+0.2\mathbf z).
\]
Unmatched masks start new tracks. The temporal window uses input frame
indices, so on SceneNN (every 20th frame), one missed update can make a
track inactive. Inactive tracks remain available for recovery, but
long-term identity consistency is not guaranteed.

\paragraph{Seeds and admission.} At query time $t$, we unite the voxels of
all masks accumulated by track $j$ up to that time and map them to the
current reconstruction $\mathcal P_t$. The largest connected component,
extracted by DBSCAN with radius 0.1\,m and minimum one sample, forms seed
$S_j$. Only seeds with $h_j\geq3$ observations and $|S_j|\geq50$ points
enter the recovery pool.

To avoid recovering existing instances, a candidate must have IoU below
0.7 with every unrefined bank instance. Candidates passing this check
undergo greedy non-maximum suppression (NMS) at IoU 0.7, in descending
order of
\[
 \frac{h_j}{h_j+2}+0.01\log(1+|S_j|).
\]
This priority determines which seeds survive NMS. Their final scores,
$0.499\,Q(S_j)$, rank them by observation consistency below all bank
instances.

The \emph{containment} control in main Table~3 and
Tables~\mbox{\ref{tab:supp-fixed-ablation}--\ref{tab:supp-components-fixed}} and~\ref{tab:supp-selection}
instead requires that the intersection with every unrefined bank instance
contain less than half of the candidate's points. This control uses native
track scores rather than ReView's confidence-based scores.

\paragraph{Observation-consistency confidence.} For an unrefined bank
instance $I$, we reuse the per-view mask matches used for refinement and
average their agreement $q_f(I)$ over all valid past views to obtain
$Q(I)$ (Algorithm~\ref{alg:review}). Unlike voting, this average includes
views with $q_f(I)<\tau_q$, retaining evidence of disagreement when
assessing reliability. Each recovery seed $S_j$ uses the same definition
for $Q(S_j)$, with its own valid views and matched masks. A valid view
without a mask contributes zero agreement; confidence is zero if no
valid views exist.

\paragraph{Implementation.} ReView uses the query-time reconstruction and
frames processed up to that time. On ScanNet200, we use every exported
frame (every 10th raw frame) with the front end's 10-pixel crop; on SceneNN,
we use every 20th posed frame. OnlineAnySeg and ReView share the same
front-end configuration across datasets and input stages. ReView requires
no additional training. The front end uses pretrained CropFormer and CLIP
and loads a frozen FCGF descriptor; CropFormer receives BGR images and
CLIP receives RGB crops.

Query-time revisiting runs on the CPU. Visibility follows the front end's
implementation, projecting points with the frame pose and depth intrinsics.
A visible point must project inside the image, have camera depth in
$(0.1,5)$\,m, and differ from the corresponding depth measurement by at most
0.08\,m. Nonzero CropFormer labels define lifted masks; label 0 is excluded.
Connectivity cleanup uses a k-d tree to find point pairs within 0.05\,m
and applies the component-size rule in Algorithm~\ref{alg:review}, without
reading images or GT.

Timing uses a tracker integrated into the front-end pass, whereas the
segmentation results use a separate implementation of the same tracking
rule. The two produce identical seeds on 311 of 324 scans; replacing the
seeds changes no reported table entry by more than 0.012 percentage points.

\paragraph{Hyperparameter selection.} ReView's default hyperparameters were
selected on 12 ScanNet200 \emph{training} scans and kept fixed across both
datasets and all input stages. The threshold for the containment rule,
used only as a recovery control, was selected separately on validation scans.
This rule is not part of ReView; selecting its threshold on validation scans
can therefore only favour the control, not ReView.

\FloatBarrier
\section{Evaluation Protocol}
\label{sec:supp-evaluation}
\paragraph{Input stages.} For $F$ processed frames on ScanNet200 or posed
frames on SceneNN, the zero-based checkpoint at fraction
$\alpha\in\{0.25,0.5,0.75,1\}$ is
\[
 t_\alpha=\max(0,\min(F-1,\operatorname{round}(\alpha F)-1)).
\]
SceneNN integrates and segments every 20th posed frame, excluding unposed
tail frames. At every stage, evaluation uses the complete GT rather than
cropping it to the observed region, so early scores also reflect incomplete
scene coverage. These stage-wise results are not directly comparable with
those in the OnlineAnySeg paper.

\paragraph{Mask mapping and evaluation protocols.} In both protocols, the
GT cloud is aligned to the query-time reconstruction using evaluation-only
ICP. Each GT vertex takes the labels of its nearest reconstruction point
within 0.15\,m, or remains unlabelled if no point is found. The search set
depends on the protocol, as described below. A predicted mask is retained
if its mapped-to-original point ratio is at least 0.75 or its mapped size
exceeds 100 points. The evaluator then applies its valid-category rule and
100-vertex minimum. An unmatched prediction counts as a false positive
only if its fraction on ignored regions does not exceed the IoU threshold.
Class-agnostic evaluation merges valid categories but preserves instance
identities; matching requires IoU strictly greater than the threshold.
GT-based operations are used only for evaluation.

The main results use OnlineAnySeg's released script
(\texttt{eval/evaluate\_seqs.py}). It searches only reconstruction points
covered by predictions, averages AP over IoU 0.50--0.90 in steps of 0.05
(the OnlineAnySeg paper states 0.50--0.95), and uses ScanNet200 category IDs on both
datasets. As a control, we evaluate our baseline reruns and ReView with a
fixed-correspondence protocol: the search includes all reconstruction
points, AP averages IoU 0.50--0.95 in steps of 0.05, and SceneNN uses NYU40
category IDs. ICP, distance limits, mask filtering and matching rules are
unchanged. For a given reconstruction, the fixed mapping is independent of
predicted masks; the two protocols also differ in threshold range and
SceneNN category IDs, so their AP difference is not a mapping-only effect.

Table~\ref{tab:supp-protocol} compares the protocols on the same saved
predictions. Starting from OnlineAnySeg's fixed-protocol AP of 14.37 on
ScanNet200, changing only the threshold range gives 15.95, changing only
the mapping gives 17.20, and changing both gives 18.96. Baseline-specific
effects are discussed in Sec.~\ref{sec:supp-results}.

\begin{table*}[t]
  \caption{\textbf{Full-input results under two evaluation protocols.}
Each method uses the same predictions under both protocols.
Fixed and Official script differ in point correspondence, IoU thresholds
for AP, and valid category IDs on SceneNN (Sec.~\ref{sec:supp-evaluation}).
Bold: best per column within each protocol.}
  \label{tab:supp-protocol}
\centering
\small
\begin{tabular}{ll|ccc|ccc}
\specialrule{1pt}{0pt}{0pt}
 \textbf{Protocol} & \textbf{Method} & \multicolumn{3}{c|}{\textbf{ScanNet200}} & \multicolumn{3}{c}{\textbf{SceneNN}}\\
& & AP & AP$_{50}$ & AP$_{25}$ & AP & AP$_{50}$ & AP$_{25}$\\
\specialrule{1pt}{0pt}{0pt}
\multirow{3}{*}{Fixed} & OnlineAnySeg & 14.37 & 34.33 & 54.90 & 16.14 & 37.60 & 58.63 \\
 & \textbf{ReView} (ours) & \textbf{17.61} & \textbf{39.43} & 61.78 & \textbf{19.02} & 40.99 & 62.87 \\
 & ReView + conf. (ours) & 15.93 & 38.71 & \textbf{62.48} & 17.43 & \textbf{41.54} & \textbf{62.95} \\
\hline
\multirow{3}{*}{Official script} & OnlineAnySeg & 18.96 & 36.77 & 55.63 & 18.12 & 36.17 & 59.86 \\
 & \textbf{ReView} (ours) & \textbf{21.38} & \textbf{41.05} & 62.51 & \textbf{20.75} & 39.33 & \textbf{62.43} \\
 & ReView + conf. (ours) & 19.34 & 40.31 & \textbf{63.12} & 18.97 & \textbf{39.82} & 62.06 \\
\specialrule{1pt}{0pt}{0pt}
\end{tabular}

\end{table*}

\paragraph{AP integration and tied scores.}
\label{sec:supp-eval}
At one IoU threshold, group predictions by confidence and include these
groups in decreasing score order. Let $(r_k,p_k)$, $k=1,\ldots,K$, be
the resulting recall--precision points, with $(r_0,p_0)=(0,1)$.
The ScanNet evaluator uses trapezoidal integration:
\begin{equation*}
 \mathrm{AP}_{\mathrm{trap}}=
 \sum_{k=1}^{K}\frac{r_k-r_{k-1}}{2}(p_k+p_{k-1}).
\end{equation*}
Step integration is $\sum_{k=1}^{K}(r_k-r_{k-1})p_k$. COCO-style AP (C-AP) averages
the precision envelope $\tilde p(r)=\max_{1\leq k\leq K:\,r_k\geq r}p_k$ at
$r\in\{0,0.01,\dots,1\}$, using 0 when no prediction-derived point reaches
$r$. For a given prediction and scoring configuration, all three use the
same TP/FP labels and differ only in integration.

When all scores are tied, $K=1$ gives one point $(R,P)$, where $R$ is
recall and $P$ is precision. ScanNet AP is then $R(1+P)/2$, step AP is
$RP$, and C-AP is
$P\cdot(\lfloor100R\rfloor+1)/101\approx RP$.

\paragraph{Low-scored recovery predictions.} With fixed correspondence and
unchanged bank masks and scores, consider appending predictions whose
scores are all below the bank's. At an IoU threshold $\theta\geq0.5$,
strict IoU matching prevents one prediction from matching two disjoint GT
instances. For predictions matching the same GT instance, the evaluator
retains the higher-scored one as TP. Appending lower-scored predictions
therefore preserves the bank's TP/FP labels and precision--recall points.
Additional curve points have nondecreasing recall, so trapezoidal and
step AP cannot decrease; neither can the COCO precision envelope.
A new TP need not strictly
increase C-AP because recall is sampled at 101 levels.
This guarantee excludes AP$_{25}$ and prediction-dependent correspondence,
which can alter mapped bank masks. Equal-budget controls
(Table~\ref{tab:supp-selection}) distinguish candidate supply from
selection quality.

\FloatBarrier
\section{Baselines}
\label{sec:supp-baselines}
We describe the configurations used to run the released implementations.
Evaluation protocols are defined in Sec.~\ref{sec:supp-evaluation}.

\paragraph{OnlineAnySeg.} We use official commit \texttt{152466e} with the
shared CropFormer masks and CLIP features (Sec.~\ref{sec:supp-method}).
Full-mesh and GT loading are disabled, and the loader receives only
current-frame poses, so inference uses only information available so far.

\paragraph{EmbodiedSAM.} Both datasets use the released checkpoint
\texttt{ESAM\_CA\_online\_epoch\_128} and the authors' preprocessing.
SAM ViT-H masks are generated every 40th raw frame (every 4th exported
ScanNet frame), followed by XYZ+RGB background clustering and nonzero-depth
handling. Each processed frame contributes 20{,}000 sampled points.
Evaluation uses EmbodiedSAM's own point cloud. On SceneNN, fixed
correspondence maps GT directly to these points after the shared
prediction-independent ICP alignment.
We also evaluate the released ESAM-E\_CA checkpoint with FastSAM masks on
every exported ScanNet200 frame; Table~\ref{tab:supp-esam-variants} compares
this configuration with the main one.

\begin{table*}[t]
  \caption{\textbf{Full-input results for two EmbodiedSAM configurations on
ScanNet200.} Ref.: view-consensus refinement; cleanup: connectivity cleanup.
Coverage: fraction of the method's own points covered by predictions,
averaged over scans. All points, 9 thr.: correspondence to all points,
with AP over IoU 0.50--0.90 in steps of 0.05; Fixed uses IoU 0.50--0.95.}
  \label{tab:supp-esam-variants}
\centering
\small
\setlength{\tabcolsep}{4pt}
\begin{tabular}{l|ccc|c|cc|cc}
\specialrule{1pt}{0pt}{0pt}
& \multicolumn{3}{c|}{\textbf{Official script}} & Fixed & \multicolumn{2}{c|}{\textbf{+ ref. + cleanup}} & & All points,\\
 \textbf{Configuration} & AP & AP$_{50}$ & AP$_{25}$ & AP & Official AP & Fixed AP & Coverage & 9 thr.\ AP\\
\specialrule{1pt}{0pt}{0pt}
ESAM\_CA + SAM ViT-H (main) & 14.80 & 36.57 & 64.43 & 20.27 & 16.67 & 18.00 & 61\% & 22.45\\
ESAM-E\_CA + FastSAM & 14.57 & 37.50 & 65.10 & 18.34 & 16.93 & 17.35 & 56\% & 20.34\\
\specialrule{1pt}{0pt}{0pt}
\end{tabular}

\end{table*}

\paragraph{SAM3D.} We use SAM3D's mask merging without its optional ensemble
with 3D over-segmentation, from official commit
\texttt{2aaa8c9}, with the original \texttt{get\_pcd} function, 0.05\,m
voxelization, the 50-point small-group rule and deterministic representative
selection. All nonzero-depth points are retained. Inputs are SAM ViT-H
masks from every exported ScanNet200 frame with a valid pose, or every
20th SceneNN frame. Each stage is run independently using only frames
available up to that point. All output instances receive score 1.

\paragraph{Stream3D.} We use official commit \texttt{8667167} on all 312
ScanNet200 scans with the shared CropFormer masks at stride 10. SceneNN
uses the same masks and every 20th posed frame as OnlineAnySeg, retaining
the ScanNet configuration. Depth is truncated at 20\,m, with no further
subsampling of the inputs. The implementation loads the complete scene
mesh before processing frames and predicts on its vertices; it is
therefore used only as a full-input reference, not for causal stage-wise
comparisons. Refinement and cleanup (main Table~4) operate on these mesh
vertices, using the same frames, poses and masks as OnlineAnySeg for
visibility, voting and confidence.

\paragraph{SceneNN cohort.} The twelve scenes are 005, 011, 015, 030, 054,
080, 089, 093, 096, 243, 263 and 322, as in OnlineAnySeg. Predictions are
pooled across scenes with scene-qualified identifiers.

\FloatBarrier
\section{Additional Results}
\label{sec:supp-results}
Unless stated otherwise, conclusions in the main text refer to
OnlineAnySeg's released evaluation script. Under fixed correspondence,
EmbodiedSAM has higher full-input AP than ReView on ScanNet200
(Table~\ref{tab:supp-fixed-main}), and refinement with cleanup lowers
EmbodiedSAM's AP at every stage (Table~\ref{tab:supp-fixed-plugin}).

\paragraph{Results with fixed correspondence.}
To assess performance under an alternative evaluation protocol, we
reevaluate the same predictions using the fixed correspondence protocol
in Sec.~\ref{sec:supp-evaluation}
(Tables~\ref{tab:supp-fixed-main} and~\ref{tab:supp-fixed-prefix}).
At full input, default ReView improves on OnlineAnySeg by 3.24 and
2.88 AP points on ScanNet200 and SceneNN, respectively. It also exceeds
this front end in every reported metric at all four input stages.
ReView has the highest full-input AP among the compared online zero-shot
methods; relative to EmbodiedSAM, its AP is 2.66 points lower on
ScanNet200 and 0.72 points higher on SceneNN.

\begin{table*}[t]
  \caption{\textbf{Full-input results with fixed correspondence.}
The predictions are identical to the corresponding rows in main Table~1;
AP uses IoU thresholds 0.50--0.95. $^{\dagger}$: SAM3D without its optional over-segmentation ensemble;
$^{\S}$: complete-mesh input, full-input reference only.
Conf.: optional bank confidence ranking.
\textbf{Bold}/\underline{underline}: best/second best per column.}
  \label{tab:supp-fixed-main}
\centering
\small
\setlength{\tabcolsep}{4pt}
\begin{tabular}{l|cc|ccc|ccc}
\specialrule{1pt}{0pt}{0pt}
\multirow{2}{*}{\textbf{Method}} & \multirow{2}{*}{\textbf{Online}} & \multirow{2}{*}{\textbf{Zero-shot}} & \multicolumn{3}{c|}{\textbf{ScanNet200}} & \multicolumn{3}{c}{\textbf{SceneNN}}\\
& & & AP & AP$_{50}$ & AP$_{25}$ & AP & AP$_{50}$ & AP$_{25}$\\
\specialrule{1pt}{0pt}{0pt}
EmbodiedSAM & \tabyes & \tabno & \textbf{20.27} & \textbf{42.19} & \textbf{63.73} & \underline{18.30} & 40.51 & 62.11 \\
SAM3D$^{\dagger}$ & \tabyes & \tabyes & 8.68 & 23.75 & 49.50 & 9.57 & 24.26 & 47.12 \\
Stream3D$^{\S}$ & \tabyes & \tabyes & 13.40 & 31.78 & 52.94 & 8.29 & 23.85 & 43.65 \\
OnlineAnySeg & \tabyes & \tabyes & 14.37 & 34.33 & 54.90 & 16.14 & 37.60 & 58.63 \\
\hline
\textbf{ReView} (ours) & \tabyes & \tabyes & \underline{17.61} & \underline{39.43} & 61.78 & \textbf{19.02} & \underline{40.99} & \underline{62.87} \\
\textbf{ReView} + conf. (ours) & \tabyes & \tabyes & 15.93 & 38.71 & \underline{62.48} & 17.43 & \textbf{41.54} & \textbf{62.95} \\
\specialrule{1pt}{0pt}{0pt}
\end{tabular}

\end{table*}

\begin{table*}[t]
  \caption{\textbf{Results during scanning with fixed correspondence.}
Percentages indicate the fraction of the input sequence processed.
C-AP$_{50}$ denotes C-AP at IoU 0.5.
\textbf{Bold}/\underline{underline}: best/second best per column within each dataset.}
  \label{tab:supp-fixed-prefix}
\centering
\footnotesize
\setlength{\tabcolsep}{3.0pt}
\begin{tabular}{ll|ccc|ccc|ccc|ccc}
\specialrule{1pt}{0pt}{0pt}
& & \multicolumn{3}{c|}{\textbf{25\%}} & \multicolumn{3}{c|}{\textbf{50\%}} & \multicolumn{3}{c|}{\textbf{75\%}} & \multicolumn{3}{c}{\textbf{100\%}}\\
\textbf{Dataset}& \textbf{Method} & AP & AP$_{50}$ & C-AP$_{50}$ & AP & AP$_{50}$ & C-AP$_{50}$ & AP & AP$_{50}$ & C-AP$_{50}$ & AP & AP$_{50}$ & C-AP$_{50}$\\
\specialrule{1pt}{0pt}{0pt}
\multirow{3}{*}{ScanNet200} & OnlineAnySeg & \underline{4.72} & \underline{12.53} & 7.10 & \underline{8.64} & 22.38 & 13.65 & 11.97 & 29.58 & 18.54 & 14.37 & 34.33 & 22.32 \\
 & \textbf{ReView} (ours) & \textbf{5.91} & \textbf{14.85} & \underline{9.47} & \textbf{10.72} & \textbf{25.63} & \underline{16.82} & \textbf{14.69} & \textbf{34.00} & \underline{22.98} & \textbf{17.61} & \textbf{39.43} & \underline{27.32} \\
 & ReView + conf. (ours) & 3.90 & 11.73 & \textbf{12.46} & 8.22 & \underline{22.63} & \textbf{22.98} & \underline{12.58} & \underline{32.13} & \textbf{32.42} & \underline{15.93} & \underline{38.71} & \textbf{38.87} \\
\hline
\multirow{3}{*}{SceneNN} & OnlineAnySeg & \underline{3.60} & \underline{8.05} & 3.52 & \underline{7.57} & \underline{18.17} & 9.36 & \underline{11.69} & 27.62 & 15.31 & 16.14 & 37.60 & 22.73 \\
 & \textbf{ReView} (ours) & \textbf{4.18} & \textbf{9.26} & \underline{4.69} & \textbf{8.91} & \textbf{20.58} & \underline{11.60} & \textbf{13.70} & \textbf{31.61} & \underline{18.93} & \textbf{19.02} & \underline{40.99} & \underline{26.17} \\
 & ReView + conf. (ours) & 2.70 & 6.58 & \textbf{7.49} & 6.08 & 16.22 & \textbf{17.21} & 10.43 & \underline{27.93} & \textbf{29.03} & \underline{17.43} & \textbf{41.54} & \textbf{42.37} \\
\specialrule{1pt}{0pt}{0pt}
\end{tabular}

\end{table*}

\begin{table}[!tb]
  \caption{\textbf{Full-input ablations and recovery controls with fixed correspondence.}
Full method: ReView; bank scores remain 1 in all configurations.
\emph{w/o}: removal of the named component.
\emph{w/o $Q$-based seed ranking} and containment-rule recovery
use native track scores for recovery seeds. Bold: best per column.}
  \label{tab:supp-fixed-ablation}
\centering
\small
\setlength{\tabcolsep}{4pt}
\begin{tabular}{l|cc|cc}
\specialrule{1pt}{0pt}{0pt}
\multirow{2}{*}{\textbf{Variant}} & \multicolumn{2}{c|}{\textbf{ScanNet200}} & \multicolumn{2}{c}{\textbf{SceneNN}}\\
& AP & AP$_{50}$ & AP & AP$_{50}$\\
\specialrule{1pt}{0pt}{0pt}
\emph{w/o} view-consensus refinement & 14.96 & 37.20 & 17.24 & 40.73 \\
\emph{w/o} connectivity cleanup & 17.01 & 38.96 & 17.90 & 39.91 \\
\emph{w/o} refinement \& cleanup & 15.40 & 37.91 & 17.08 & 40.46 \\
\emph{w/o} track recovery & 16.61 & 36.02 & 18.16 & 38.22 \\
\emph{w/o} $Q$-based seed ranking & 17.46 & 39.02 & 18.96 & 40.95 \\
containment-rule recovery & 17.16 & 37.87 & 18.47 & 39.25 \\
\hline
\textbf{Full method} & \textbf{17.61} & \textbf{39.43} & \textbf{19.02} & \textbf{40.99} \\
\specialrule{1pt}{0pt}{0pt}
\end{tabular}

\end{table}

\paragraph{Component combinations.}
Tables~\ref{tab:supp-components-official} and~\ref{tab:supp-components-fixed}
evaluate individual components and their combinations.
Combining refinement and cleanup yields higher AP than either alone
on both datasets under both evaluation protocols, supporting their
complementary roles. Under fixed correspondence, the full method
outperforms component-removal and alternative recovery configurations
in AP and AP$_{50}$ (Table~\ref{tab:supp-fixed-ablation}).
Cleanup improves both metrics on both datasets at full input, but lowers
some metrics at intermediate stages (Table~\ref{tab:supp-cleanup-prefix}).

\begin{table*}[t]
  \caption{\textbf{Full-input component combinations with the official script.}
\tabyes/--: enabled/disabled. Recovery: cont.\ uses containment admission
and native track scores; IoU uses IoU$<$0.7 admission and seed scores
$0.499\,Q$. Both use track-priority NMS. Conf.: bank confidence ranking;
C-AP$_{50}$ denotes C-AP at IoU 0.5. Bold: best per column.}
  \label{tab:supp-components-official}
\centering
\small
\setlength{\tabcolsep}{3.5pt}
\begin{tabular}{l|cccc|ccc|ccc}
\specialrule{1pt}{0pt}{0pt}
\multirow{2}{*}{\textbf{Variant}} & \multirow{2}{*}{\textbf{Refinement}} & \multirow{2}{*}{\textbf{Cleanup}} & \multirow{2}{*}{\textbf{Recovery}} & \multirow{2}{*}{\textbf{Conf.}} & \multicolumn{3}{c|}{\textbf{ScanNet200}} & \multicolumn{3}{c}{\textbf{SceneNN}}\\
 &  &  &  &  & AP & AP$_{50}$ & C-AP$_{50}$ & AP & AP$_{50}$ & C-AP$_{50}$ \\
\specialrule{1pt}{0pt}{0pt}
Front end only & -- & -- & -- & -- & 18.96 & 36.77 & 24.50 & 18.12 & 36.17 & 21.52 \\
 & \tabyes & -- & -- & -- & 18.67 & 36.71 & 24.32 & 19.10 & 37.62 & 22.78 \\
 & -- & \tabyes & -- & -- & 18.68 & 36.32 & 24.26 & 18.15 & 35.35 & 20.70 \\
 & \tabyes & \tabyes & -- & -- & 20.36 & 38.06 & 26.22 & 19.65 & 36.68 & 21.72 \\
 & -- & -- & cont. & -- & 19.93 & 39.08 & 26.82 & 19.16 & 38.69 & 23.70 \\
 & \tabyes & -- & cont. & -- & 19.63 & 38.78 & 26.56 & 19.75 & 38.27 & 23.60 \\
 & \tabyes & -- & IoU & -- & 19.76 & 39.80 & 27.66 & 19.61 & 37.74 & 23.12 \\
\textbf{Full method} & \tabyes & \tabyes & IoU & -- & \textbf{21.38} & \textbf{41.05} & 29.08 & \textbf{20.75} & 39.33 & 24.56 \\
\hline
 & -- & -- & -- & \tabyes & 16.78 & 35.67 & 35.74 & 15.57 & 35.31 & 36.37 \\
 & -- & -- & IoU & \tabyes & 18.43 & 39.91 & 40.19 & 17.56 & \textbf{40.57} & \textbf{41.42} \\
 & \tabyes & -- & IoU & \tabyes & 18.08 & 39.04 & 39.40 & 18.03 & 38.63 & 39.39 \\
\textbf{Full method} + conf. & \tabyes & \tabyes & IoU & \tabyes & 19.34 & 40.31 & \textbf{40.64} & 18.97 & 39.82 & 40.59 \\
\specialrule{1pt}{0pt}{0pt}
\end{tabular}

\end{table*}

\begin{table*}[t]
  \caption{\textbf{Full-input component combinations with fixed correspondence.}
Configurations and notation follow Table~\ref{tab:supp-components-official}.
Bold: best per column.}
  \label{tab:supp-components-fixed}
\centering
\small
\setlength{\tabcolsep}{3.5pt}
\begin{tabular}{l|cccc|ccc|ccc}
\specialrule{1pt}{0pt}{0pt}
\multirow{2}{*}{\textbf{Variant}} & \multirow{2}{*}{\textbf{Refinement}} & \multirow{2}{*}{\textbf{Cleanup}} & \multirow{2}{*}{\textbf{Recovery}} & \multirow{2}{*}{\textbf{Conf.}} & \multicolumn{3}{c|}{\textbf{ScanNet200}} & \multicolumn{3}{c}{\textbf{SceneNN}}\\
 &  &  &  &  & AP & AP$_{50}$ & C-AP$_{50}$ & AP & AP$_{50}$ & C-AP$_{50}$ \\
\specialrule{1pt}{0pt}{0pt}
Front end only & -- & -- & -- & -- & 14.37 & 34.33 & 22.32 & 16.14 & 37.60 & 22.73 \\
 & \tabyes & -- & -- & -- & 15.98 & 35.46 & 23.24 & 17.03 & 36.96 & 22.20 \\
 & -- & \tabyes & -- & -- & 13.95 & 33.68 & 21.54 & 16.32 & 37.86 & 22.85 \\
 & \tabyes & \tabyes & -- & -- & 16.61 & 36.02 & 24.25 & 18.16 & 38.22 & 23.61 \\
 & -- & -- & cont. & -- & 14.88 & 36.11 & 24.03 & 16.43 & 38.61 & 23.66 \\
 & \tabyes & -- & cont. & -- & 16.52 & 37.28 & 25.02 & 17.32 & 37.97 & 23.42 \\
 & \tabyes & -- & IoU & -- & 17.01 & 38.96 & 26.70 & 17.90 & 39.91 & 25.16 \\
\textbf{Full method} & \tabyes & \tabyes & IoU & -- & \textbf{17.61} & \textbf{39.43} & 27.32 & \textbf{19.02} & 40.99 & 26.17 \\
\hline
 & -- & -- & -- & \tabyes & 13.10 & 33.75 & 34.09 & 14.57 & 37.86 & 38.73 \\
 & -- & -- & IoU & \tabyes & 14.13 & 37.33 & 37.51 & 15.51 & 40.72 & 41.61 \\
 & \tabyes & -- & IoU & \tabyes & 15.58 & 38.27 & 38.47 & 16.56 & 40.69 & 41.49 \\
\textbf{Full method} + conf. & \tabyes & \tabyes & IoU & \tabyes & 15.93 & 38.71 & \textbf{38.87} & 17.43 & \textbf{41.54} & \textbf{42.37} \\
\specialrule{1pt}{0pt}{0pt}
\end{tabular}

\end{table*}

\begin{table}[!tb]
  \caption{\textbf{Connectivity cleanup across input stages.}
Fixed correspondence is used. Both variants retain refinement and
confidence-scored recovery seeds, with bank scores fixed at 1.
Bold: better within each pair.}
  \label{tab:supp-cleanup-prefix}
\centering
\footnotesize
\setlength{\tabcolsep}{3pt}
\begin{tabular}{ll|cc|cc}
\specialrule{1pt}{0pt}{0pt}
 \textbf{Stage} & \textbf{Output} & \multicolumn{2}{c|}{\textbf{ScanNet200}} & \multicolumn{2}{c}{\textbf{SceneNN}}\\
& & AP & AP$_{50}$ & AP & AP$_{50}$\\
\specialrule{1pt}{0pt}{0pt}
\multirow{2}{*}{25\%} & w/o cleanup & 5.78 & 14.78 & \textbf{4.27} & \textbf{10.13} \\
 & \textbf{ReView} & \textbf{5.91} & \textbf{14.85} & 4.18 & 9.26 \\
\hline
\multirow{2}{*}{50\%} & w/o cleanup & 10.41 & \textbf{25.74} & 8.65 & \textbf{21.04} \\
 & \textbf{ReView} & \textbf{10.72} & 25.63 & \textbf{8.91} & 20.58 \\
\hline
\multirow{2}{*}{75\%} & w/o cleanup & 14.37 & \textbf{34.25} & 13.28 & \textbf{32.28} \\
 & \textbf{ReView} & \textbf{14.69} & 34.00 & \textbf{13.70} & 31.61 \\
\hline
\multirow{2}{*}{100\%} & w/o cleanup & 17.01 & 38.96 & 17.90 & 39.91 \\
 & \textbf{ReView} & \textbf{17.61} & \textbf{39.43} & \textbf{19.02} & \textbf{40.99} \\
\specialrule{1pt}{0pt}{0pt}
\end{tabular}

\end{table}

\begin{figure}[!tb]
\centering
\includegraphics[width=\columnwidth]{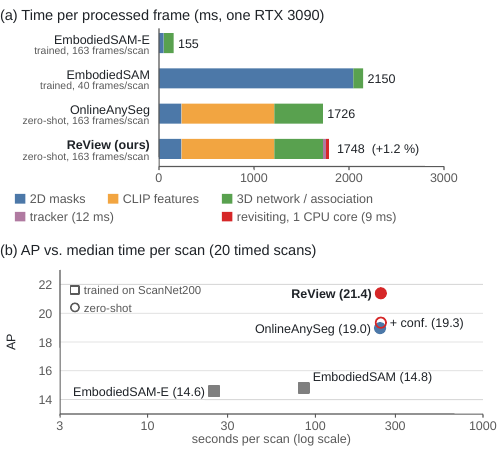}
\caption{\textbf{Efficiency.} Timing uses 20 ScanNet200 scans and one RTX
3090; revisiting uses one CPU core. (a)~Time per processed frame, with one
query per scan amortized over its frames. (b)~Full-input AP on all 312 validation scans
versus median time per scan over the 20 timed scans (log scale); squares: trained on ScanNet200,
circles: zero-shot. ESAM-E uses network timings sampled every 4th frame;
its per-scan time is estimated from per-frame time and frame count.
Archived-frame loading is excluded from query timing.}
\label{fig:supp-eff}
\end{figure}

\paragraph{Other online methods.} Tables~\ref{tab:supp-fixed-plugin}
and~\ref{tab:supp-plugin-prefix} extend main Table~4 to both evaluation
protocols and available input stages; Stream3D is evaluated only at
full input. Refinement with cleanup improves SAM3D's AP at every stage
and Stream3D's full-input AP under both protocols. For EmbodiedSAM, the
combination improves AP at 75--100\% input under the official script,
leaves it nearly unchanged at 50\%, and lowers it at 25\%; with fixed
correspondence, AP decreases at every stage. Thus, gains across front
ends depend on the base method and evaluation protocol.

\begin{table*}[t]
  \caption{\textbf{Results on other front ends with fixed correspondence.}
Evaluation uses ScanNet200 at each input stage.
Conf.: observation-consistency ranking; other configurations retain
original instance scores. $^{\dagger}$: SAM3D without its optional over-segmentation ensemble; $^{\S}$: complete-mesh
input, evaluated only at full input. Bold: best per column within each method.}
  \label{tab:supp-fixed-plugin}
\centering
\small
\begin{tabular}{l|cc|cc|cc|cc}
\specialrule{1pt}{0pt}{0pt}
& \multicolumn{2}{c|}{\textbf{25\%}} & \multicolumn{2}{c|}{\textbf{50\%}} & \multicolumn{2}{c|}{\textbf{75\%}} & \multicolumn{2}{c}{\textbf{100\%}}\\
\textbf{Method} & AP & AP$_{50}$ & AP & AP$_{50}$ & AP & AP$_{50}$ & AP & AP$_{50}$\\
\specialrule{1pt}{0pt}{0pt}
SAM3D$^{\dagger}$ & 3.92 & 11.23 & 6.25 & 17.56 & 7.97 & 22.17 & 8.68 & 23.75 \\
\quad + refinement & 5.46 & \textbf{13.35} & 9.05 & 21.51 & 11.61 & 26.63 & 13.18 & 29.60 \\
\quad + refinement + cleanup & \textbf{5.56} & 13.27 & \textbf{9.51} & \textbf{21.63} & \textbf{12.21} & 26.93 & \textbf{14.03} & 30.32 \\
\quad + refinement + cleanup + conf. & 3.66 & 10.78 & 7.26 & 19.61 & 10.67 & \textbf{26.97} & 12.84 & \textbf{31.17} \\
\hline
Stream3D$^{\S}$ & -- & -- & -- & -- & -- & -- & 13.40 & 31.78 \\
\quad + refinement & -- & -- & -- & -- & -- & -- & 14.62 & 33.42 \\
\quad + refinement + cleanup & -- & -- & -- & -- & -- & -- & \textbf{16.44} & \textbf{35.19} \\
\quad + refinement + cleanup + conf. & -- & -- & -- & -- & -- & -- & 14.89 & 34.84 \\
\hline
EmbodiedSAM & \textbf{8.46} & \textbf{19.68} & \textbf{14.07} & \textbf{31.16} & \textbf{18.01} & \textbf{38.56} & \textbf{20.27} & \textbf{42.19} \\
\quad + refinement & 5.95 & 16.95 & 10.52 & 28.23 & 14.11 & 36.18 & 16.58 & 40.75 \\
\quad + refinement + cleanup & 6.18 & 16.87 & 11.21 & 28.50 & 15.14 & 36.62 & 18.00 & 41.97 \\
\specialrule{1pt}{0pt}{0pt}
\end{tabular}

\end{table*}

\begin{table*}[t]
  \caption{\textbf{Results on other front ends with the official script.}
Evaluation uses ScanNet200 at each input stage, with configurations
and notation as in Table~\ref{tab:supp-fixed-plugin}.
Bold: best per column within each method.}
  \label{tab:supp-plugin-prefix}
\centering
\small
\begin{tabular}{l|cc|cc|cc|cc}
\specialrule{1pt}{0pt}{0pt}
& \multicolumn{2}{c|}{\textbf{25\%}} & \multicolumn{2}{c|}{\textbf{50\%}} & \multicolumn{2}{c|}{\textbf{75\%}} & \multicolumn{2}{c}{\textbf{100\%}}\\
\textbf{Method} & AP & AP$_{50}$ & AP & AP$_{50}$ & AP & AP$_{50}$ & AP & AP$_{50}$\\
\specialrule{1pt}{0pt}{0pt}
SAM3D$^{\dagger}$ & 4.33 & 11.23 & 6.92 & 17.56 & 8.83 & 22.12 & 9.63 & 23.75 \\
\quad + refinement & 6.13 & 13.54 & 10.18 & 21.85 & 13.15 & 27.14 & 14.93 & 30.14 \\
\quad + refinement + cleanup & \textbf{6.75} & \textbf{14.02} & \textbf{11.40} & \textbf{22.75} & \textbf{14.69} & 28.42 & \textbf{16.81} & 31.60 \\
\quad + refinement + cleanup + conf. & 4.47 & 11.55 & 8.66 & 20.87 & 12.68 & \textbf{28.43} & 15.24 & \textbf{32.52} \\
\hline
Stream3D$^{\S}$ & -- & -- & -- & -- & -- & -- & 19.18 & 34.31 \\
\quad + refinement & -- & -- & -- & -- & -- & -- & 17.60 & 34.56 \\
\quad + refinement + cleanup & -- & -- & -- & -- & -- & -- & \textbf{20.96} & \textbf{36.92} \\
\quad + refinement + cleanup + conf. & -- & -- & -- & -- & -- & -- & 18.88 & 36.59 \\
\hline
EmbodiedSAM & \textbf{6.74} & \textbf{17.25} & 10.93 & 27.39 & 13.55 & 33.81 & 14.80 & 36.57 \\
\quad + refinement & 3.17 & 11.56 & 5.46 & 19.63 & 7.17 & 25.25 & 8.40 & 28.24 \\
\quad + refinement + cleanup & 6.16 & 16.80 & \textbf{11.02} & \textbf{28.23} & \textbf{14.44} & \textbf{35.80} & \textbf{16.67} & \textbf{40.00} \\
\specialrule{1pt}{0pt}{0pt}
\end{tabular}

\end{table*}

\paragraph{Hyperparameter sensitivity.} Table~\ref{tab:supp-sensitivity} gives
the numerical results for main Fig.~6. Past-view fractions of 50\%,
25\% and 12.5\% retain every second, fourth and eighth processed frame,
respectively, for both voting and confidence computation. The default
uses all processed frames; front-end inputs remain unchanged.
On both datasets, all tested settings exceed front-end AP under both
evaluation protocols.

\begin{table*}[t]
  \caption{\textbf{Hyperparameter sensitivity at full input.}
One hyperparameter is varied at a time; all others retain their defaults.
The first four numerical columns report AP and AP$_{50}$ with the
official script; the last two report AP with fixed correspondence.}
  \label{tab:supp-sensitivity}
\centering
\small
\setlength{\tabcolsep}{5pt}
\begin{tabular}{l|cc|cc|cc}
\specialrule{1pt}{0pt}{0pt}
& \multicolumn{2}{c|}{\textbf{ScanNet200}} & \multicolumn{2}{c|}{\textbf{SceneNN}} & \multicolumn{2}{c}{\textbf{Fixed corr.\ AP}}\\
\textbf{Setting} & AP & AP$_{50}$ & AP & AP$_{50}$ & ScanNet200 & SceneNN\\
\specialrule{1pt}{0pt}{0pt}
\textbf{Defaults} & 21.38 & 41.05 & 20.75 & 39.33 & 17.61 & 19.02 \\
\hline
Vote threshold on $q_f$: 0.2 & 21.26 & 40.65 & 20.76 & 38.84 & 17.57 & 19.04 \\
\quad 0.4 & 21.51 & 41.37 & 20.71 & 39.05 & 17.65 & 18.91 \\
\quad 0.5 & 21.49 & 41.32 & 21.19 & 40.09 & 17.57 & 19.38 \\
Majority ratio: 0.4 & 21.69 & 41.21 & 21.15 & 39.52 & 18.11 & 19.47 \\
\quad 0.6 & 20.71 & 40.26 & 19.87 & 38.64 & 16.24 & 17.97 \\
Min.\ observations per point: 1 & 21.36 & 40.82 & 20.74 & 39.67 & 17.81 & 19.21 \\
\quad 3 & 21.42 & 41.22 & 20.92 & 39.27 & 17.33 & 19.01 \\
Past views (voting and confidence): 50\% & 21.27 & 40.97 & 20.84 & 38.63 & 17.15 & 18.61 \\
\quad 25\% & 20.94 & 40.67 & 20.12 & 38.48 & 16.41 & 18.16 \\
\quad 12.5\% & 20.37 & 40.15 & 19.62 & 38.10 & 15.56 & 17.33 \\
Cleanup radius: 0.03\,m & 20.80 & 40.35 & 20.80 & 39.02 & 16.74 & 18.91 \\
\quad 0.08\,m & 21.69 & 41.60 & 20.80 & 38.81 & 18.16 & 19.01 \\
Cleanup keep fraction: 0.3 & 21.36 & 40.99 & 21.00 & 39.86 & 17.61 & 19.24 \\
\quad 0.7 & 21.20 & 40.63 & 20.79 & 39.33 & 17.61 & 19.04 \\
Recovery IoU gate: 0.5 & 21.52 & 41.07 & 20.96 & 39.08 & 17.53 & 18.92 \\
\quad 0.9 & 21.08 & 40.60 & 20.76 & 39.11 & 17.65 & 19.14 \\
\hline
OnlineAnySeg & 18.96 & 36.77 & 18.12 & 36.17 & 14.37 & 16.14 \\
\specialrule{1pt}{0pt}{0pt}
\end{tabular}

\end{table*}

\paragraph{AP integration and scoring controls.}
To distinguish the effects of scoring and integration,
Table~\ref{tab:supp-metrics} compares ScanNet AP, step AP and C-AP
under fixed correspondence, including an EmbodiedSAM control with all
scores set to 1. For each scoring configuration, the three integrations
use the same matches. With instance masks unchanged, ReView's confidence
ranking increases step AP and C-AP but decreases ScanNet AP,
showing that the effect of scoring depends on the integration rule.

\begin{table*}[!t]
  \caption{\textbf{AP integration and scoring controls at full input.}
With fixed correspondence, ScanNet AP (trapezoidal), step AP and C-AP
use the same matches within each row. The all-scores-1
control changes only EmbodiedSAM's scores; ReView + conf.\ changes only
bank scores, keeping instance masks and recovery scores fixed.}
  \label{tab:supp-metrics}
\centering
\small
\begin{tabular}{ll|ccc|ccc|ccc}
\specialrule{1pt}{0pt}{0pt}
& & \multicolumn{3}{c|}{\textbf{ScanNet}} & \multicolumn{3}{c|}{\textbf{Step}} & \multicolumn{3}{c}{\textbf{C-AP}}\\
\textbf{Dataset} &  \textbf{Method} & AP & AP$_{50}$ & AP$_{25}$ & AP & AP$_{50}$ & AP$_{25}$ & AP & AP$_{50}$ & AP$_{25}$\\
\specialrule{1pt}{0pt}{0pt}
\multirow{5}{*}{\shortstack[l]{ScanNet200\\100\%}} & EmbodiedSAM & 20.27 & 42.19 & 63.73 & 20.27 & 42.20 & 63.73 & 20.59 & 42.43 & 63.88\\
 & EmbodiedSAM (all scores 1) & 17.78 & 36.14 & 56.91 & 7.99 & 21.12 & 42.77 & 8.00 & 21.25 & 42.90\\
 & OnlineAnySeg & 14.37 & 34.33 & 54.90 & 7.29 & 22.25 & 44.92 & 7.33 & 22.32 & 44.55\\
 &  \textbf{ReView} (ours) & 17.61 & 39.43 & 61.78 & 9.75 & 27.42 & 51.73 & 9.77 & 27.32 & 51.68\\
 &  ReView + conf. (ours) & 15.93 & 38.71 & 62.48 & 15.94 & 38.72 & 62.49 & 16.17 & 38.87 & 62.65\\
\hline
\multirow{5}{*}{\shortstack[l]{SceneNN\\100\%}} & EmbodiedSAM & 18.30 & 40.51 & 62.11 & 18.35 & 40.53 & 62.14 & 18.83 & 40.75 & 62.01\\
 & EmbodiedSAM (all scores 1) & 14.69 & 30.87 & 52.05 & 4.49 & 13.12 & 32.13 & 4.50 & 13.09 & 31.82\\
 & OnlineAnySeg & 16.14 & 37.60 & 58.63 & 7.37 & 22.73 & 44.74 & 7.39 & 22.73 & 44.58\\
 &  \textbf{ReView} (ours) & 19.02 & 40.99 & 62.87 & 9.33 & 26.20 & 49.09 & 9.36 & 26.17 & 48.92\\
 &  ReView + conf. (ours) & 17.43 & 41.54 & 62.95 & 17.55 & 41.65 & 63.05 & 18.18 & 42.37 & 63.76\\
\specialrule{1pt}{0pt}{0pt}
\end{tabular}

\end{table*}

\paragraph{Simple confidence baselines.} To test whether observation-consistency
confidence $Q$ merely reflects instance size or observation count, we keep all
instance masks and recovery scores fixed and replace only the bank scores by the
final point count $|\hat I|$, the number of valid views $|\mathcal G_I|$, or a
random value, each mapped monotonically to $[0.5,1)$ so that bank instances still
rank above recovery seeds (Table~\ref{tab:supp-conf-baselines}). $Q$ yields the
highest C-AP$_{50}$ on both datasets (40.64 and 40.59), about 7 points above the
best simple score (point count: 33.67 and 33.05), followed by the number of
valid views (31.54 and 32.19). $Q$ also lowers ScanNet AP the least: the two
simple scores reduce it by 7.2--8.4 points, whereas $Q$ reduces it by
1.8--2.0 points. Fixed correspondence gives the same ordering.

\begin{table*}[t]
  \caption{\textbf{Bank confidence baselines at full input} (official script).
  Masks and recovery scores are identical in all rows; only bank scores change.
  Random: mean over five seeds. Bold: best per column.}
  \label{tab:supp-conf-baselines}
  \centering\small
  \begin{tabular}{l|ccc|ccc}
  \toprule
  & \multicolumn{3}{c|}{ScanNet200} & \multicolumn{3}{c}{SceneNN}\\
  Bank score & AP & AP$_{50}$ & C-AP$_{50}$ & AP & AP$_{50}$ & C-AP$_{50}$\\
  \midrule
  All 1 (ReView) & \textbf{21.38} & \textbf{41.05} & 29.08 & \textbf{20.75} & 39.33 & 24.56\\
  $Q$ (ReView + conf.) & 19.34 & 40.31 & \textbf{40.64} & 18.97 & \textbf{39.82} & \textbf{40.59}\\
  Point count $|\hat I|$ & 13.46 & 32.01 & 33.67 & 13.56 & 31.39 & 33.05\\
  Valid views $|\mathcal G_I|$ & 12.97 & 30.86 & 31.54 & 12.63 & 31.43 & 32.19\\
  Random & 12.22 & 29.32 & 29.83 & 10.23 & 24.85 & 26.25\\
  \bottomrule
  \end{tabular}
\end{table*}

\paragraph{Scene-level comparisons.} Table~\ref{tab:supp-contrasts-official}
compares ReView with OnlineAnySeg at each input stage under both evaluation
protocols. We obtain 95\% percentile intervals from 10{,}000 paired bootstrap
resamples: rooms for ScanNet200 (142 rooms, retaining all scans in each
sampled room) and scenes for SceneNN (12 scenes). Means are scan-weighted;
percentiles use linear interpolation. On ScanNet200, the intervals for
mean AP gains lie above zero at every stage under both protocols.
SceneNN also has positive mean AP gains throughout, but the intervals
include zero at early stages under the official script.

\begin{table*}[!t]
  \caption{\textbf{Scan-level differences: default ReView minus OnlineAnySeg.}
Bank scores are fixed at 1, without optional confidence ranking.
$\Delta$AP and $\Delta$C-AP$_{50}$ are scan-weighted mean differences in
percentage points, not differences in pooled AP. Brackets give 95\%
bootstrap confidence intervals; W/T/L denotes scan-level wins/ties/losses.}
  \label{tab:supp-contrasts-official}
\centering
\footnotesize
\begin{tabular}{ll|cc|cc}
\specialrule{1pt}{0pt}{0pt}
 & & \multicolumn{2}{c|}{\textbf{$\Delta$AP}} & \multicolumn{2}{c}{\textbf{$\Delta$C-AP$_{50}$}}\\
\textbf{Dataset} & \textbf{Stage} & \textbf{Mean [95\% CI]} & \textbf{W/T/L} & \textbf{Mean [95\% CI]} & \textbf{W/T/L}\\
\specialrule{1pt}{0pt}{0pt}
\multicolumn{6}{l}{\textbf{(a) Official script}}\\
\hline
\multirow{4}{*}{ScanNet200} & 25\% & 1.07 [0.76, 1.39] & 224/5/83 & 2.89 [2.28, 3.50] & 208/54/50\\
 & 50\% & 1.74 [1.30, 2.26] & 238/0/74 & 3.49 [2.72, 4.29] & 224/23/65\\
 & 75\% & 2.26 [1.71, 2.82] & 242/2/68 & 4.56 [3.64, 5.45] & 233/19/60\\
 & 100\% & 2.53 [1.92, 3.10] & 245/1/66 & 4.99 [3.90, 6.06] & 239/18/55\\
\hline
\multirow{4}{*}{SceneNN} & 25\% & 0.68 [-0.22, 1.70] & 8/1/3 & 0.81 [-1.98, 3.57] & 7/2/3\\
 & 50\% & 1.27 [-0.06, 2.70] & 9/0/3 & -0.12 [-2.49, 2.06] & 7/2/3\\
 & 75\% & 2.51 [0.44, 4.75] & 9/0/3 & 3.36 [-0.46, 7.62] & 8/0/4\\
 & 100\% & 3.13 [0.95, 5.29] & 9/0/3 & 3.05 [-0.90, 6.76] & 9/0/3\\
\specialrule{1pt}{0pt}{0pt}
\multicolumn{6}{l}{\textbf{(b) Fixed correspondence}}\\
\hline
\multirow{4}{*}{ScanNet200} & 25\% & 1.43 [1.09, 1.85] & 251/5/56 & 3.17 [2.44, 4.07] & 209/59/44\\
 & 50\% & 2.57 [2.08, 3.15] & 270/0/42 & 4.25 [3.36, 5.25] & 229/21/62\\
 & 75\% & 3.33 [2.72, 4.05] & 270/0/42 & 5.60 [4.57, 6.68] & 240/17/55\\
 & 100\% & 3.67 [3.11, 4.28] & 273/0/39 & 5.82 [4.91, 6.78] & 248/15/49\\
\hline
\multirow{4}{*}{SceneNN} & 25\% & 0.73 [0.16, 1.43] & 10/0/2 & 1.09 [-1.04, 2.97] & 8/2/2\\
 & 50\% & 1.79 [0.49, 3.16] & 10/0/2 & 2.75 [-0.50, 6.35] & 8/1/3\\
 & 75\% & 2.78 [0.76, 5.08] & 10/0/2 & 5.07 [0.45, 10.34] & 8/0/4\\
 & 100\% & 3.62 [1.33, 5.95] & 10/0/2 & 4.21 [-0.40, 9.45] & 8/0/4\\
\specialrule{1pt}{0pt}{0pt}
\end{tabular}

\label{tab:supp-contrasts-ours}
\end{table*}

\paragraph{Recovery at an equal budget.} Table~\ref{tab:supp-selection}
compares rules at their native candidate counts and at an equal
per-scene budget to separate candidate supply from admission quality.
The controls share the same track pool, unrefined OnlineAnySeg bank,
fixed correspondence and native track scores; selection uses no ground
truth. This experiment isolates recovery selection rather than
evaluating the full ReView configuration.

Truncated to the containment rule's per-scene count, the IoU$<$0.7 and
ungated selectors change scene-mean AP$_{50}$ relative to the containment
rule by $+0.02$ ($[-0.08,0.11]$) and $-0.27$ ($[-0.41,-0.14]$) points on
ScanNet200 and by $+0.46$ ($[0.19,0.73]$) and $+0.26$ ($[0.10,0.42]$) on
SceneNN. At equal counts, the IoU gate produces fewer duplicate false
positives than ungated selection on both datasets, while containment
suppresses duplicates more strongly. Retaining the larger, untruncated
IoU$<$0.7 pool further increases AP$_{50}$, indicating that recovery
gains reflect both candidate coverage and duplicate control.

\begin{table*}[!t]
  \caption{\textbf{Recovery candidate counts and admission rules.}
Fixed correspondence, the unrefined OnlineAnySeg bank and native track
scores are used. AP$_{50}$ uses ScanNet integration;
$\#$ counts appended candidates, and Dup counts duplicate false positives
in the output at IoU 0.5. Equal $K$: each scene uses the containment
rule's candidate count.}
  \label{tab:supp-selection}
\centering
\small
\begin{tabular}{l|cccc|cccc}
\specialrule{1pt}{0pt}{0pt}
\multirow{2}{*}{\textbf{Appended to OnlineAnySeg}} & \multicolumn{4}{c|}{\textbf{ScanNet200}} & \multicolumn{4}{c}{\textbf{SceneNN}}\\
& AP$_{50}$ & C-AP$_{50}$ & $\#$ & Dup & AP$_{50}$ & C-AP$_{50}$ & $\#$ & Dup\\
\specialrule{1pt}{0pt}{0pt}
None & 34.33 & 22.32 & 0 & 5 & 37.60 & 22.73 & 0 & 0 \\
Containment $<$0.5 & 36.11 & 24.03 & 7,966 & 54 & 38.61 & 23.66 & 638 & 8 \\
\hline
IoU$<$0.7 & 37.44 & 25.24 & 32,176 & 1,021 & 40.50 & 25.55 & 2,967 & 114 \\
No bank gate & 37.05 & 24.91 & 43,015 & 5,076 & 40.21 & 25.25 & 3,475 & 256 \\
IoU$<$0.7, equal $K$ & 36.27 & 24.27 & 7,966 & 454 & 39.64 & 24.75 & 638 & 57 \\
No bank gate, equal $K$ & 35.48 & 23.42 & 7,966 & 2,393 & 39.19 & 24.27 & 638 & 134 \\
\specialrule{1pt}{0pt}{0pt}
\end{tabular}

\end{table*}

\paragraph{Efficiency.} We time 20 randomly drawn ScanNet200 scans (seed
2027) on one RTX 3090 (Fig.~\ref{fig:supp-eff}). The front end takes
1726\,ms per processed frame. ReView adds an amortized 21\,ms (1.2\%):
12\,ms for tracking and 9\,ms for revisiting on one CPU core (AMD EPYC
7413). This amortization assumes one query per scan. A query takes
1.4\,s at the median and at most 3.4\,s, excluding archived-frame I/O.
Vote counters use a median of 12\,MB per scan.
If ReView instead queries every $N$ processed frames and recomputes each query
from the archived frames, the median total time per ScanNet200 scan is 7.2,
12.4, 23.4 and 112.6\,s for $N=20$, 10, 5 and 1, i.e., 2.8\%, 4.9\%, 9.3\% and
44\% of the front-end time. The archived depth maps, poses and 2D masks take a
median of 17\,MB per scan. Because each query re-accumulates votes over all
archived frames, the cost of frequent queries grows with sequence length; this
is a limitation of the current implementation.

\section{Qualitative Results}
\label{sec:supp-qual}
\begin{figure*}[!t]
\centering
\includegraphics[width=0.86\textwidth]{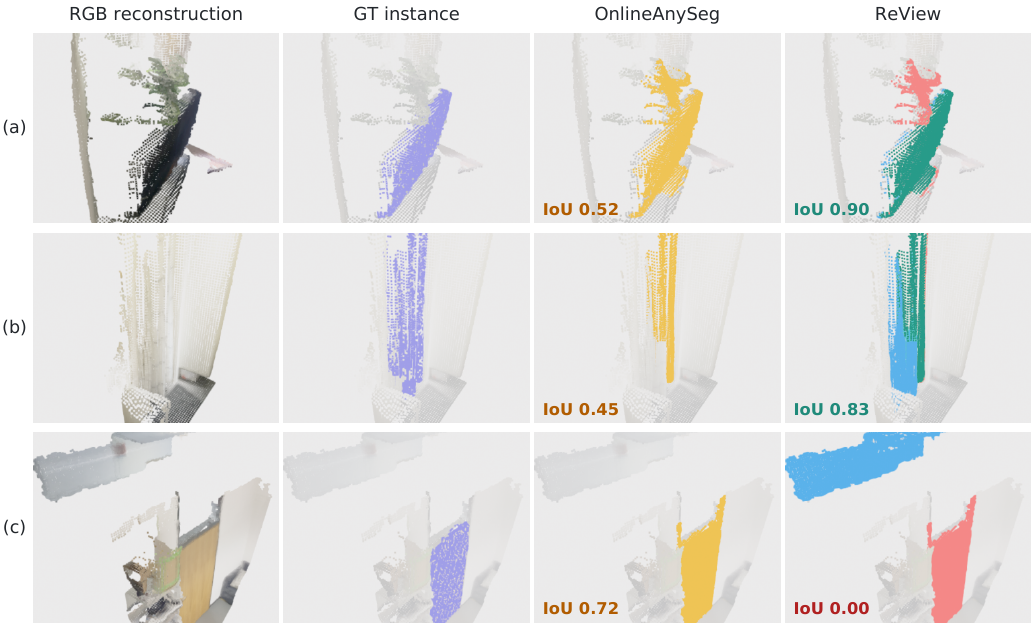}
\caption{\textbf{Qualitative results of refinement and cleanup.}
ScanNet200, full input. (a,b)~Successful cases; (c)~failure.
In the ReView column, teal/blue/red indicate kept/added/removed points;
red points are excluded from the final output. Numbers give IoU between
the predicted instance and the corresponding GT instance.}
\label{fig:supp-success}
\label{fig:supp-fail}
\end{figure*}

\paragraph{Instance close-ups.} Figure~\ref{fig:supp-success} shows two
successful cases and one failure of refinement and cleanup. ReView
removes incorrectly included plant points in (a) and adds missing points
to an existing instance in (b). From bank instances with at least
1{,}500 points in every third ScanNet200 scan, we select two of the
largest IoU gains and the largest loss. GT is used only for example
selection and evaluation.

\paragraph{Failure case.} In Fig.~\ref{fig:supp-fail}(c), matched masks
across views include a larger neighbouring surface. Voting adds it to
the instance, and cleanup retains it while removing the original target,
reducing target IoU to zero. Errors shared across views can thus persist
through voting and affect connectivity cleanup.

\paragraph{Instance-level changes.} In 22 of the 4{,}046 bank instances
(0.5\%) from the sampled scans, cleanup discards the component containing
most original points. Among the 1{,}311 instances with at least
1{,}500 points, IoU with the originally matched GT improves for 51.9\%
and worsens for 44.5\%, with a mean change near zero. These instance-level
statistics and pooled AP measure different aspects of performance;
AP gains do not imply improvements in every instance or in mean IoU.

\paragraph{Supplementary video.}
{\Urlmuskip=0mu\relax\def\UrlBreaks{\do\/\do\.}\def\UrlBigBreaks{\do\:}The video is available at \url{https://github.com/zhangyuanyang21/ReView-3DIS}.}
The video shows segmentation evolving across four scans, with GT,
OnlineAnySeg and ReView displayed side by side. Predictions are exported
at approximately 2.5\% input intervals, using only observations available
at each checkpoint. The scans comprise scene0608\_01 from main Fig.~1;
scene0702\_02 and scene0064\_00, with the largest and median AP gains
among ScanNet200 scans containing at least 10 GT instances; and
SceneNN 096, with a near-median gain. Gains are measured at full input.
For display, predicted instances successfully matched to GT by Hungarian
matching on point IoU ($\geq 0.25$) use the corresponding GT colours.
GT is used only for reference, colouring and example selection,
not for segmentation inference. ReView uses the ReView + conf.\
scores from the main text, with lighter colours indicating lower
confidence. Numbered circles highlight selected examples of corrected regions
and recovered objects. During the call-outs, the recovered object in circle 2 is
shown at full colour. The closing metrics are taken from main Tables~1--2.

\FloatBarrier

\end{document}